\documentclass[10pt,twocolumn,letterpaper]{article}

\usepackage[pagenumbers]{cvpr} 

\newcommand{\algname}{SemanticMoments}
\newcommand{\dataset}{SimMotion}
\newcommand{\datasetS}{\dataset-Synthetic}
\newcommand{\datasetR}{\dataset-Real}

\usepackage{graphicx}
\usepackage{subcaption}
\usepackage{adjustbox}
\usepackage{caption}

\definecolor{cvprblue}{rgb}{0.21,0.49,0.74}
\usepackage[pagebackref,breaklinks,colorlinks,allcolors=cvprblue]{hyperref}

\def\confName{CVPR}
\def\confYear{2026}

\title{\algname: Training-Free Motion Similarity via Third Moment Features}

\author{
  Saar Huberman$^{1,2}$ \quad 
  Kfir Goldberg$^1$ \quad 
  Or Patashnik$^2$ \quad
  Sagie Benaim$^3$ \quad
  Ron Mokady$^1$ \\
  \vspace{1em}
  {\normalsize $^1$BRIA AI \quad $^2$Tel Aviv University \quad $^3$Hebrew University of Jerusalem}
}

\begin{document}

\twocolumn[{
\renewcommand\twocolumn[1][]{#1}

\maketitle
\begin{center}
  \centering
  \vspace{-0.8cm}
    \includegraphics[width=\textwidth]{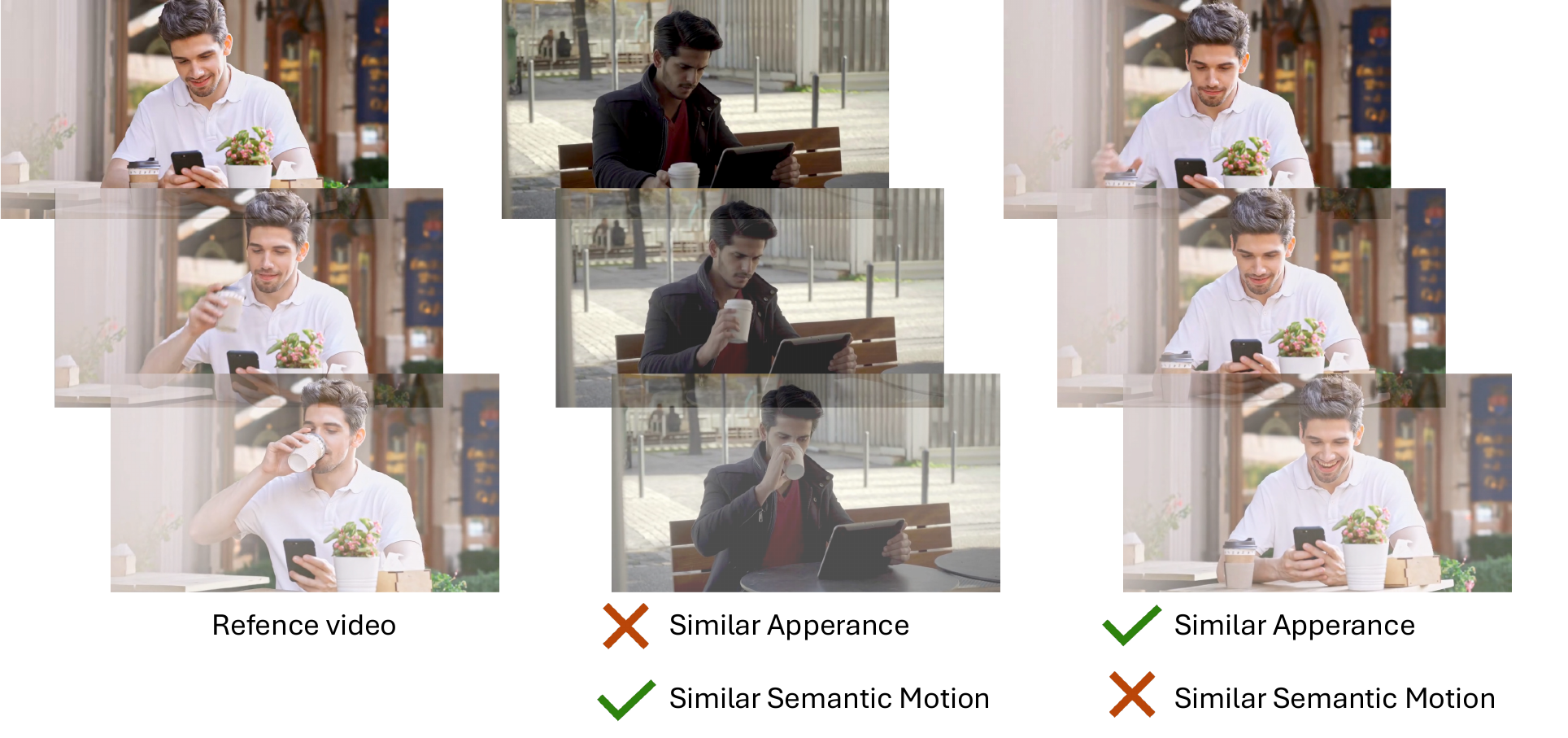}
  \captionof{figure}{\textbf{Motion-centric retrieval with Semantic Moments.}
  Existing video-similarity methods over-rely on static appearance and scene context, overlooking temporal dynamics. 
  Our approach retrieves clips that match the \emph{semantic motion}.
  We retrieve the drinking-coffee motion across identities, disentangling motion from appearance,  while all baselines similarly return look-alikes and miss the action.}
  \label{fig:teaser}
\end{center}
}]

\begin{abstract}

Retrieving videos based on semantic motion is a fundamental, yet unsolved, problem. Existing video representation approaches overly rely on static appearance and scene context rather than motion dynamics, a bias inherited from their training data and objectives. Conversely, traditional motion-centric inputs like optical flow lack the semantic grounding needed to understand high-level motion. To demonstrate this inherent bias, we introduce the SimMotion benchmarks, combining controlled synthetic data with a new human-annotated real-world dataset. We show that existing models perform poorly on these benchmarks, often failing to disentangle motion from appearance. To address this gap, we propose SemanticMoments, a simple, training-free method that computes temporal statistics (specifically, higher-order moments) over features from pre-trained semantic models. Across our benchmarks, SemanticMoments consistently outperforms existing RGB, flow, and text-supervised methods. This demonstrates that temporal statistics in a semantic feature space provide a scalable and perceptually grounded foundation for motion-centric video understanding.

\end{abstract}
\section{Introduction}

Humans perceive motion not as raw pixel displacement, but as meaningful, structured, and semantic change over time \cite{blake2007perception, giese2003metric, pollick2003visual, thurman2016revisiting, vangeneugden2009functional}. Two videos may differ visually, but convey similar motion when comparable entities undergo analogous temporal transformations (as can be seen in ~\cref{fig:teaser}). In this view, motion similarity captures how perceptual structure unfolds over time at the semantic level. 
Retrieving videos that share such similar motion is a fundamental yet largely unsolved problem, which we tackle in this paper. 
Such capability would benefit a wide range of applications, from constructing motion-centric datasets to enhancing motion control in generative video models.

We find that most existing video retrieval methods rely on representations that are biased towards static appearance and scene context rather than motion dynamics, often producing results that are visually similar but dynamically unrelated\footnote{A similar phenomenon was observed in image classification networks~\cite{geirhos2018imagenet}, where models trained on ImageNet are biased toward recognizing local texture rather than global structure.}. This bias stems from the data and objectives used for training. The problem often begins with the datasets themselves: action-recognition labels are a flawed proxy for motion, as identical actions can exhibit distinct motions while different actions can share similar dynamics. 

\begin{figure}[t]
    \centering
    \setlength{\tabcolsep}{0.0pt}
    {\small
    \begin{tabular}{c c c}
        \textbf{Kinetics} & \textbf{UCF101} & \textbf{HMDB-51} \\
        \includegraphics[width=0.33\linewidth, height=0.19\linewidth]{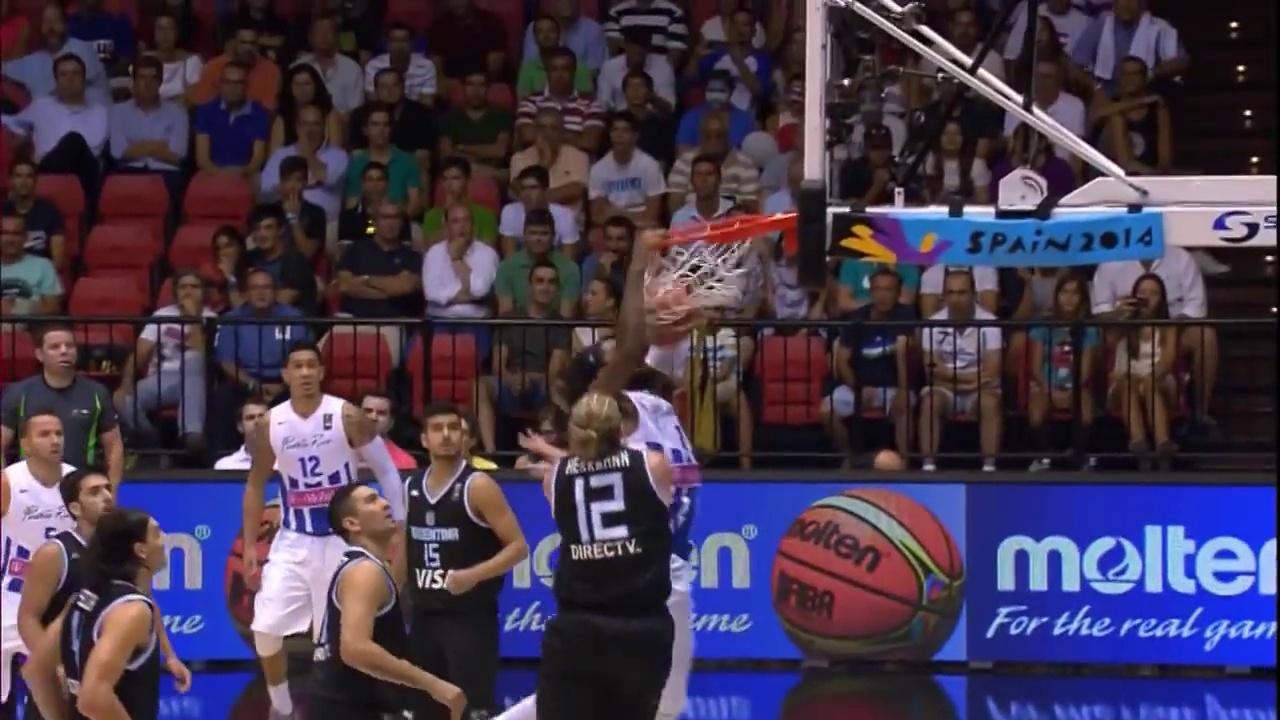} &
        \includegraphics[width=0.33\linewidth, height=0.19\linewidth]{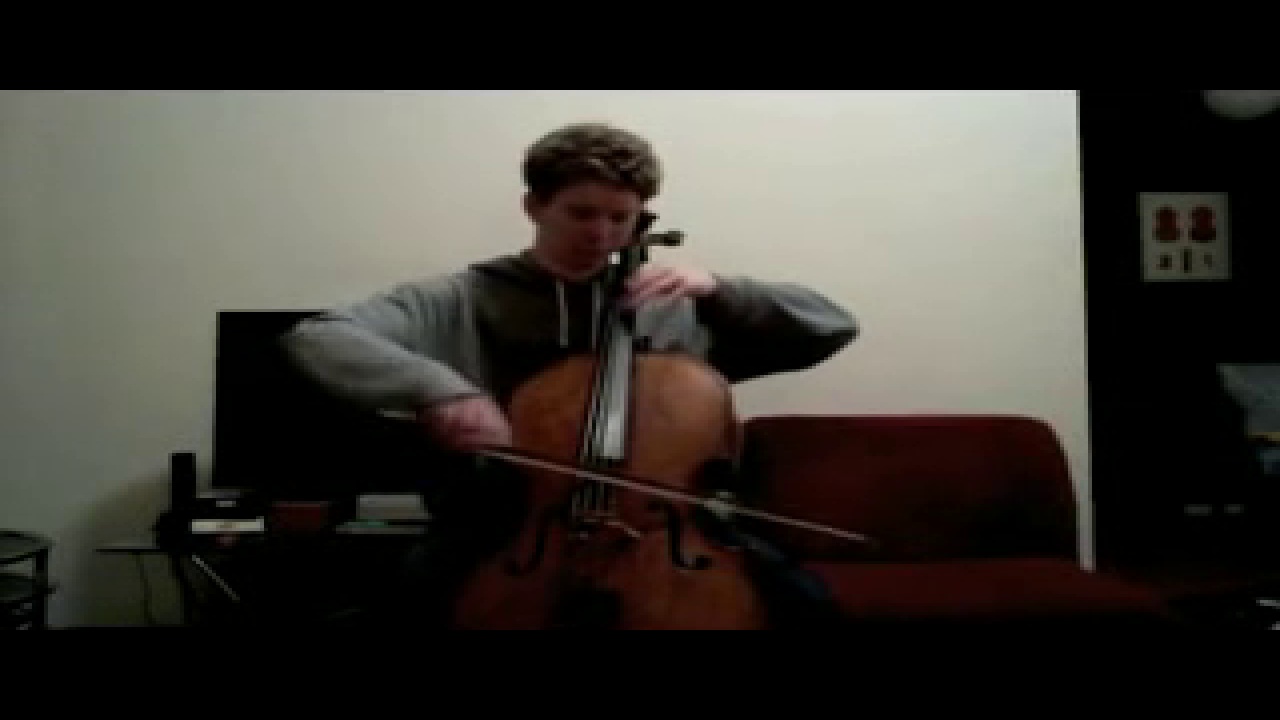} &
        \includegraphics[width=0.33\linewidth, height=0.19\linewidth]{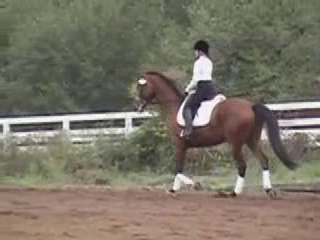}
        
        \\

        Dunking basketball & Playing Cello & Ride horse
        \\
        \includegraphics[width=0.33\linewidth, height=0.19\linewidth]{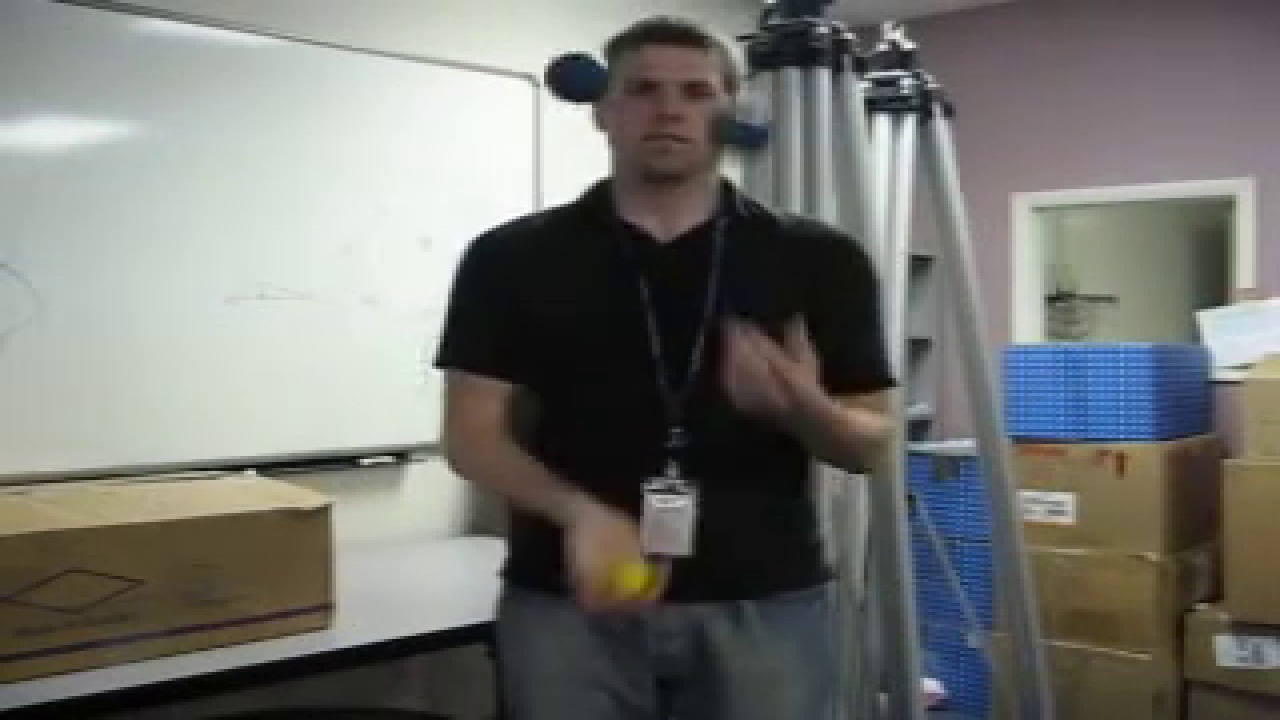} &
        \includegraphics[width=0.33\linewidth, height=0.19\linewidth]{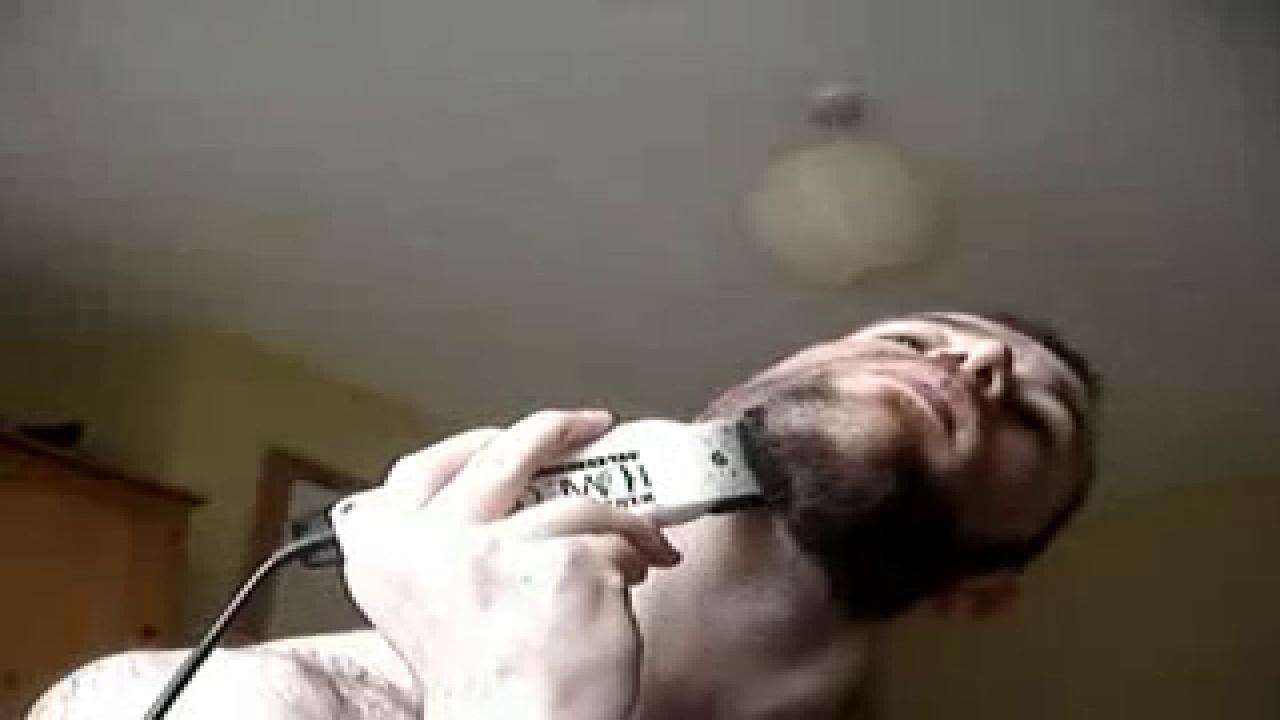} &
        \includegraphics[width=0.33\linewidth, height=0.19\linewidth]{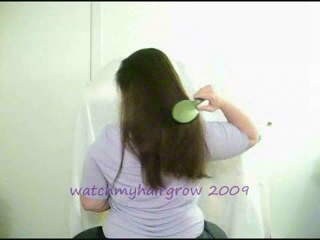}
        
        \\[2pt]

        Juggling balls & Shaving beard & Brush hair

    \end{tabular}
    }
    \vspace{-0.2cm}
    \caption{
    \textbf{Current benchmarks are appearance-centric.}
    We show random frames from popular video-retrieval datasets. In many cases, static objects (e.g., a cello, a razor) or scene context (e.g., a basketball court) suffice to identify the action label (e.g., Playing Cello, Shaving Beard) without observing motion. This bias enables high accuracy from purely appearance-based cues, discouraging models from learning true temporal dynamics.
    }
    \label{fig:single_frame}
    \vspace{-0.4cm}
\end{figure}

Consequently, models trained on such data learn to exploit these dataset biases. Supervised approaches \cite{simonyan2014two, carreira2017quo, feichtenhofer2019slowfast, gberta_2021_ICML}, for instance, can often recognize an action category from a single frame (~\cref{fig:single_frame}), learning to encode static appearance cues (e.g., background, objects, and clothing) rather than true temporal structure. We find this appearance bias is general; it also occurs in self-supervised RGB methods \cite{qian2021spatiotemporal, pan2021videomoco, dorkenwald2022scvrl, tong2022videomae, han2020self}, which often learn to prioritize simple appearance consistency over modeling complex temporal change. The primary alternative, methods relying solely on optical flow, presents the opposite problem: while robust to appearance, they are unable to capture the important semantic information that defines perceptual motion.

To illustrate this bias concretely, we introduce the \textit{SimMotion-Synthetic} dataset. This synthetic dataset consists of video pairs that share identical motion but differ in controlled factors such as viewpoint and visual style. As our retrieval experiments show, existing methods often fail to identify videos with the same motion, as their representations are sensitive to these non-motion factors. This emphasizes the need for representations that explicitly capture motion. However, collecting large-scale annotated data for such training is prohibitively expensive.

To this end, we introduce \textit{SemanticMoments}, a simple, training-free motion representation designed to move beyond static appearance cues. Our core insight is that standard temporal aggregation, such as average pooling (first statistical moment), effectively captures average appearance but discards the rich temporal dynamics of how features change. To explicitly capture this motion, \textit{SemanticMoments} computes a richer set of temporal statistics, specifically the higher-order moments (e.g., variance and skewness), over patch-level embeddings from pretrained semantic models like DINO \cite{caron2021emerging, oquab2023dinov2}. This results in a compact descriptor that summarizes high-level, structured change. It also leverages the strong semantic correspondences of DINO features, which track meaningful object parts as trajectories in the feature space \cite{tumanyan2024dino}. 
Our approach directly captures semantic dynamics and requires no optical flow, labeled data, or additional training, making it broadly compatible with off-the-shelf backbones.

To evaluate motion-based retrieval in realistic settings, we introduce \textit{SimMotion-Real}, our second benchmark, which consists of human-annotated video pairs labeled for perceptual motion similarity, independently of appearance.
Since producing such annotations is time-consuming, the dataset is intentionally small but carefully curated to capture diverse, natural motion patterns. This benchmark enables rigorous, real-world evaluation of motion representations and complements our controlled synthetic analysis.
We show that our method consistently retrieves videos with similar motion,  
 outperforming existing approaches. 
To summarize, we provide the following contributions: 
\begin{itemize}
\item We identify and analyze the dominant appearance bias in current video representations, showing they prioritize static cues over motion dynamics.
\item We introduce the \textit{SimMotion} benchmarks, a new suite of synthetic and real-world human-annotated datasets rigorously evaluating perceptual motion similarity.
\item We propose \textit{SemanticMoments}, a simple, efficient and training-free method that represents motion using the temporal statistics of semantic features.  \textit{SemanticMoments} outperforms existing state-of-the-art approaches.
\end{itemize}

\section{Related Work}

\noindent \textbf{Video Representation Learning.} \quad Prior work retrieves videos based on the similarity of features produced by some pretrained model, trained using different supervision modes: 
\textit{action recognition}, \textit{multimodal supervision}, and \textit{self-supervision}.

\textit{Action Recognition.} Early models,  trained on UCF-101~\cite{soomro2012ucf101}, HMDB-51~\cite{kuehne2011hmdb}, and Kinetics~\cite{kay2017kinetics}, learn to classify videos into predefined categories. The architectures used include Two-Stream Networks~\cite{simonyan2014two} and I3D~\cite{carreira2017quo} with parallel multi-stream inputs (RGB and optical flow), or SlowFast~\cite{feichtenhofer2019slowfast} with varying frame rates. While these architectures were designed to decouple appearance and motion, their shared training objective undermined this goal. Since both paths were trained to predict the same action labels, which are often defined by static objects or scenes, the models learned that the appearance-based path was the most reliable predictor for minimizing loss. As a result, they inherited the dataset's bias, continuing to rely on appearance cues rather than motion.

\textit{Multimodal Supervision.} Recent methods adapt models pretrained on text-image/video data via contrastive learning. CLIP4Clip~\cite{luo2022clip4clip} aggregates CLIP~\cite{radford2021learning} frame features for video-text alignment; VideoCLIP~\cite{xu2021videoclip} jointly learns audio, visual, and textual embeddings; and X-CLIP~\cite{ma2022x} uses cross-modal transformers for fine-grained frame-text interaction. Despite strong retrieval results, these models inherit a limitation of language supervision: motion is often underspecified (e.g., ``a person walking'' or ``an object rotating'' can match many distinct videos).

\textit{Self Supervision.} Self-supervised approaches learn embeddings without labels using contrastive (e.g., CVRL~\cite{qian2021spatiotemporal}, VideoMoCo~\cite{pan2021videomoco}) or masked-prediction objectives (e.g., VideoMAE~\cite{tong2022videomae}, VideoPrism~\cite{zhao2024videoprism}). While many of these methods \textit{attempt} to target motion, such as SCVRL~\cite{dorkenwald2022scvrl} (via shuffling) or the more recent V-JEPA~\cite{bardes2024revisiting, assran2025v} (by predicting latent spatiotemporal regions based on JEPA~\cite{assran2023self}), they remain sensitive to their prediction target design. Their objectives often make learning static appearance consistency the simplest path to minimizing loss. For example, in both masked and predictive modeling, a model is "incentivized to preserve appearance even as motion changes", thus inheriting the very appearance bias we aim to solve.

\noindent \textbf{Disentangling Motion from Appearance.} \quad 
Recent works seek to isolate motion, primarily for generative tasks like motion transfer. MoFT~\cite{xiao2024video} identifies motion-sensitive components in diffusion models via PCA; DIFTFlow~\cite{pondaven2025video} extracts motion trajectories from attention maps; SMM~\cite{yatim2024space} reduces appearance bias using inter-frame differences and spatial marginal means; and MotionClone~\cite{ling2024motionclone} derives sparse temporal attention to guide motion imitation. While they pursue disentanglement for generative tasks, their goal is not to produce the compact, global embeddings required for large-scale \textit{retrieval}, which is the focus of our work.

In parallel, self-supervised vision transformers such as DINO support motion reasoning: DiVE~\cite{huang2024dive} uses DINOv2 to extract localized trajectories and preserve subject identity via LoRA adapters, and MotionShot~\cite{liu2025motionshot} fuses DINO with Stable Diffusion features to align high-level semantics with low-level structure for controllable transfer. Together, these generative approaches show that pretrained visual features capture motion-relevant structure across time. We adopt this insight but shift the goal: rather than transfer, we use pretrained feature extractors to obtain a compact, training-free representation for motion-based video similarity.


\begin{figure}[t]
  \centering
  \setlength{\tabcolsep}{0.8pt}
  \resizebox{\columnwidth}{!}{%
    \begin{tabular}{ccccc}
        \LARGE Static & \LARGE Dyn-App & \LARGE Dyn-Obj &  \LARGE View & \LARGE Style \\
      \includegraphics[width=0.19\textwidth]{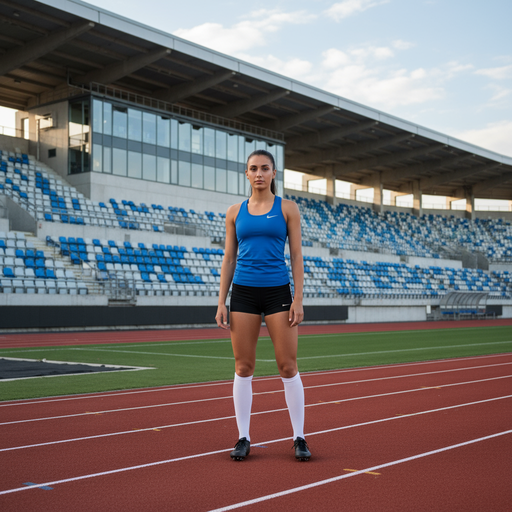} &
      \includegraphics[width=0.19\textwidth]{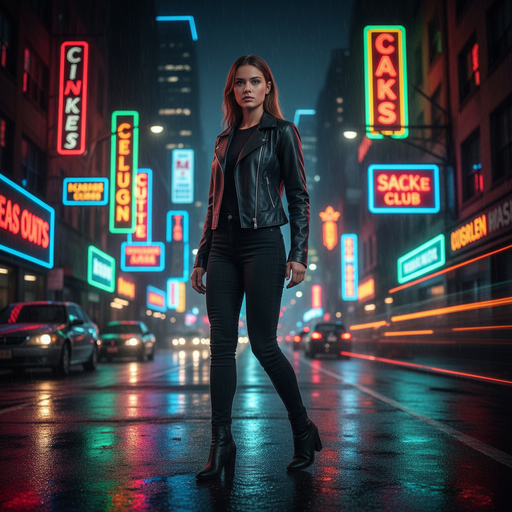} &
      \includegraphics[width=0.19\textwidth]{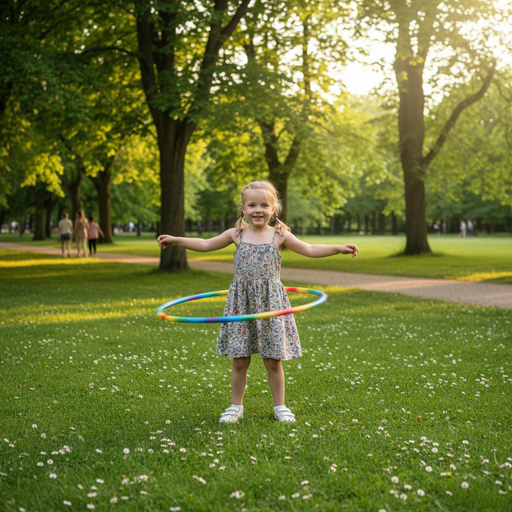} &
      \includegraphics[width=0.19\textwidth]{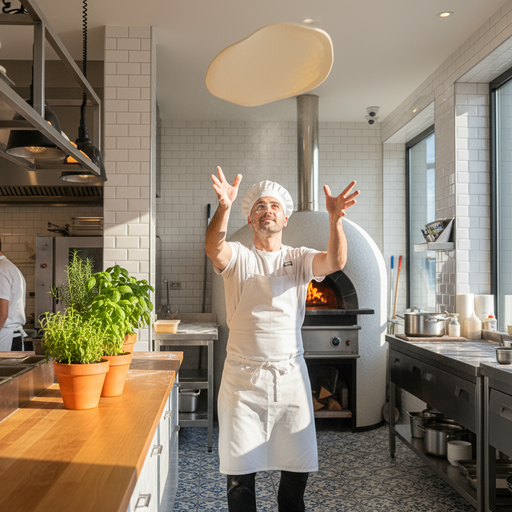} &
      \includegraphics[width=0.19\textwidth]{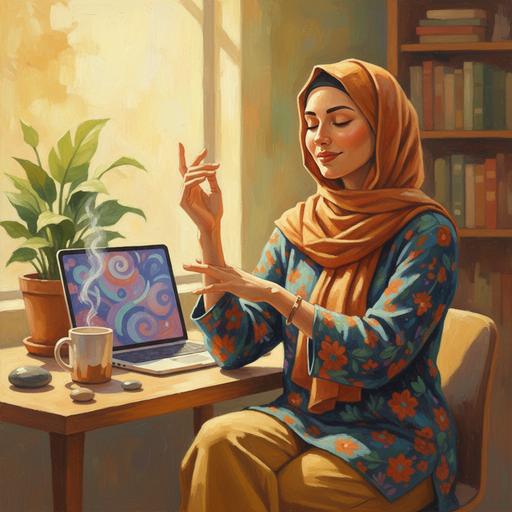} \\
      \includegraphics[width=0.19\textwidth]{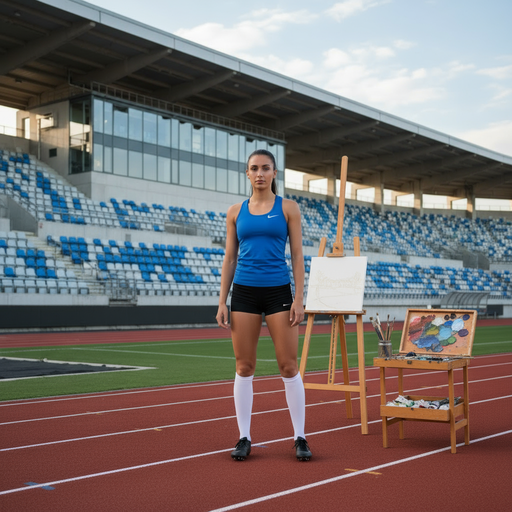} &
      \includegraphics[width=0.19\textwidth]{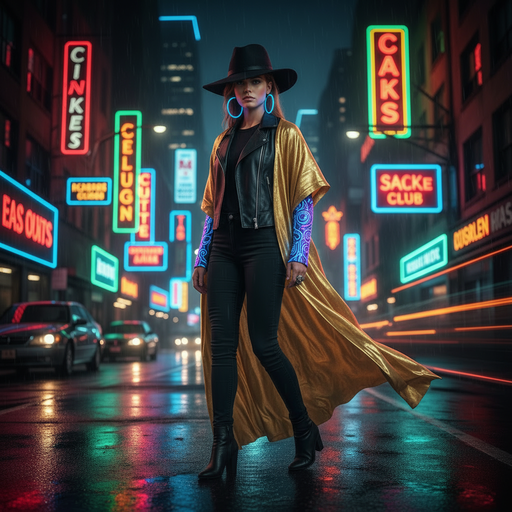} &
      \includegraphics[width=0.19\textwidth]{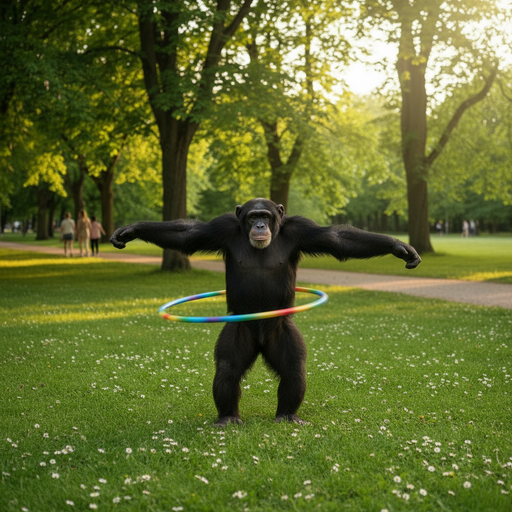} &
      \includegraphics[width=0.19\textwidth]{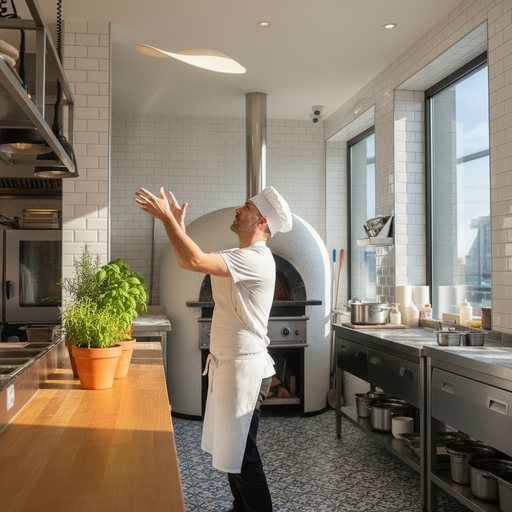} &
      \includegraphics[width=0.19\textwidth]{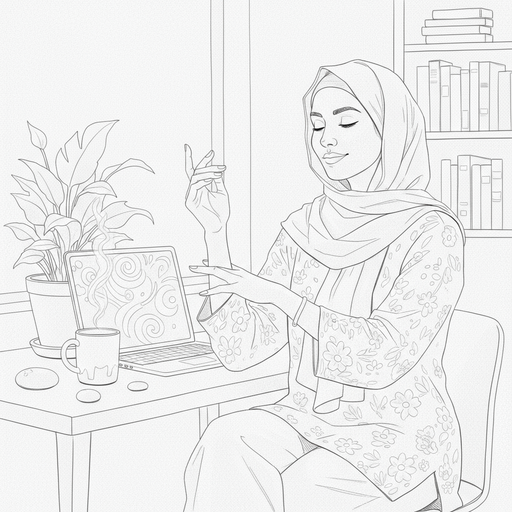} \\
    \end{tabular}%
  }
  \caption{\textbf{Controlled variation in SimMotion-Synthetic.} We visualize sample pairs from the five distinct categories in our benchmark. From left to right: \textbf{Static Object} (background varies), \textbf{Dynamic Appearance} (subject clothing/attributes vary), \textbf{Dynamic Object} (subject identity varies), \textbf{View} (camera angle varies), and \textbf{Scene Style} (rendering style varies). In each column, the top and bottom videos are temporally synchronized and share identical motion dynamics, differing only in the specified visual factor.}
  \label{fig:sync_motions_dataset}
  \vspace{-0.3cm}
\end{figure}

\section{The “\dataset” benchmarks}
\label{sec:datasets}

Existing methods for motion similarity or motion retrieval are typically evaluated on action recognition benchmarks.
However, as discussed earlier, defining motion through discrete categories or textual descriptions is inherently limited. Categories such as “walking”, “jumping”, or “dancing” provide only coarse, high-level descriptions, useful for naming an action but insufficient to convey the structure and the dynamics of the motion. In fact, as demonstrated in Fig.~\ref{fig:single_frame}, such categories can be identified from single frames only without observing motion.

Human perception of motion operates across multiple levels of abstraction: at a coarse level, we distinguish action types (e.g., walking vs. jumping), while at a finer level, we perceive variations in structure and dynamics~\cite{blake2007perception, thurman2016revisiting, pollick2003visual, giese2003metric, vangeneugden2009functional}. 
For example, the category 'dancing' (coarse level) includes both 'waltz' and 'breakdancing' (fine level), which have entirely different motion structures. Furthermore, even two different instances of a 'waltz' will vary significantly in their execution and dynamics, yet are perceptually similar.
As categorical labels capture only part of this hierarchy, measuring motion similarity should also incorporate structural and dynamic properties. Because similarity in these aspects is continuous rather than discrete, benchmarks should be based on relative similarity measures rather than categorical labels.


\begin{figure*}[t]
\centering
\setlength{\tabcolsep}{0.3pt}
{\scriptsize
\newsavebox{\Abox}
\newlength{\Aht}
\sbox{\Abox}{%
  \begin{minipage}[t]{0.48\textwidth}
    \centering
    \begin{tabular}{c c c}
        \textbf{Original} & \textbf{Pixar} & \textbf{Paintbrush}\\
        \includegraphics[width=0.33\linewidth]{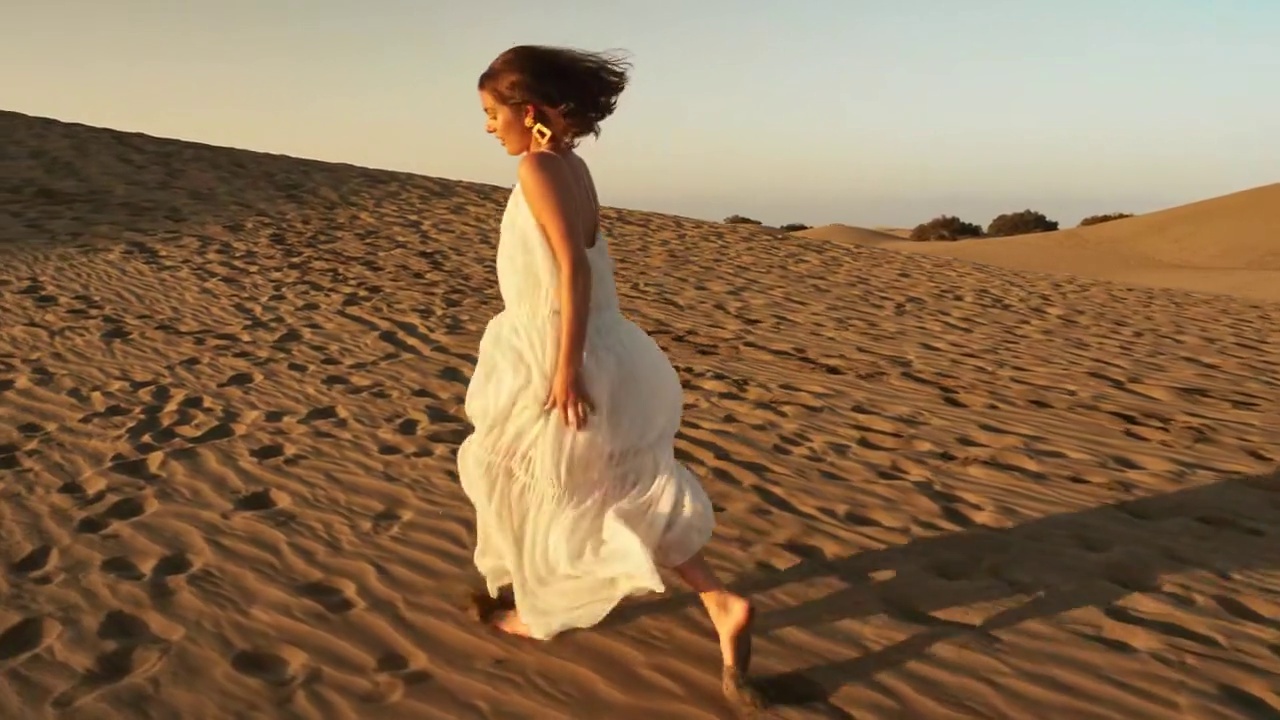} &
        \includegraphics[width=0.33\linewidth]{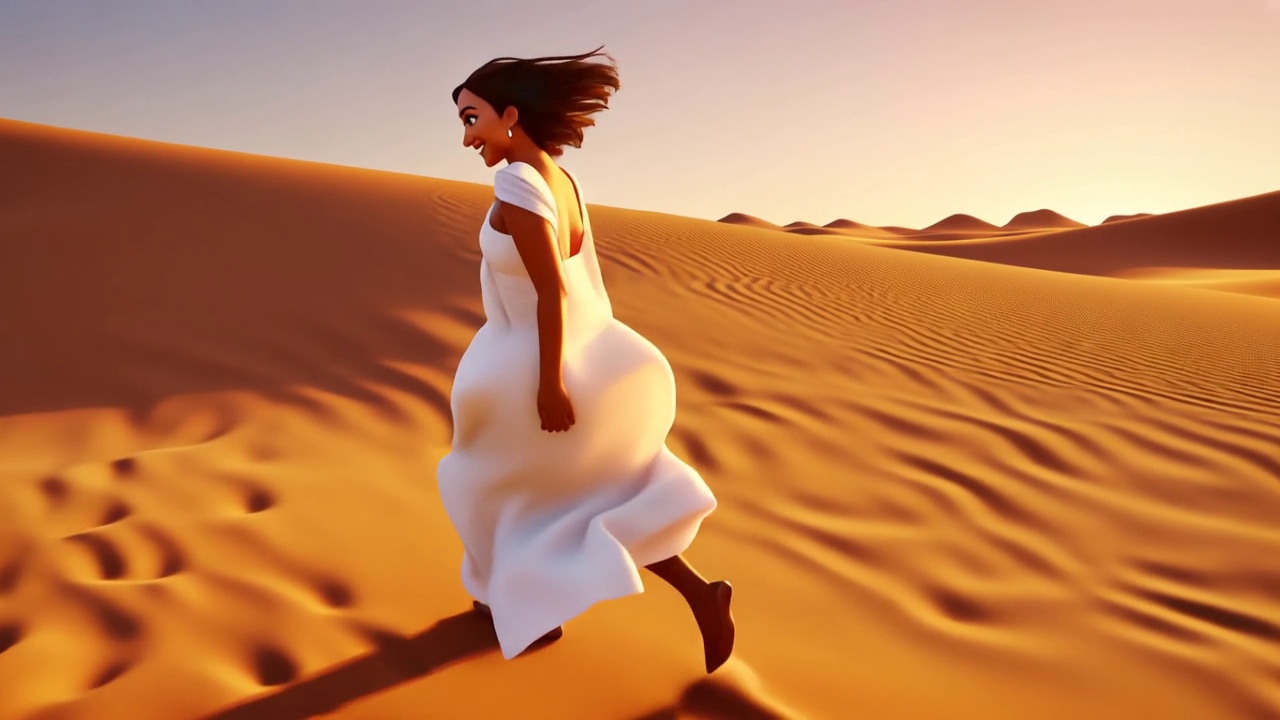} &
        \includegraphics[width=0.33\linewidth]{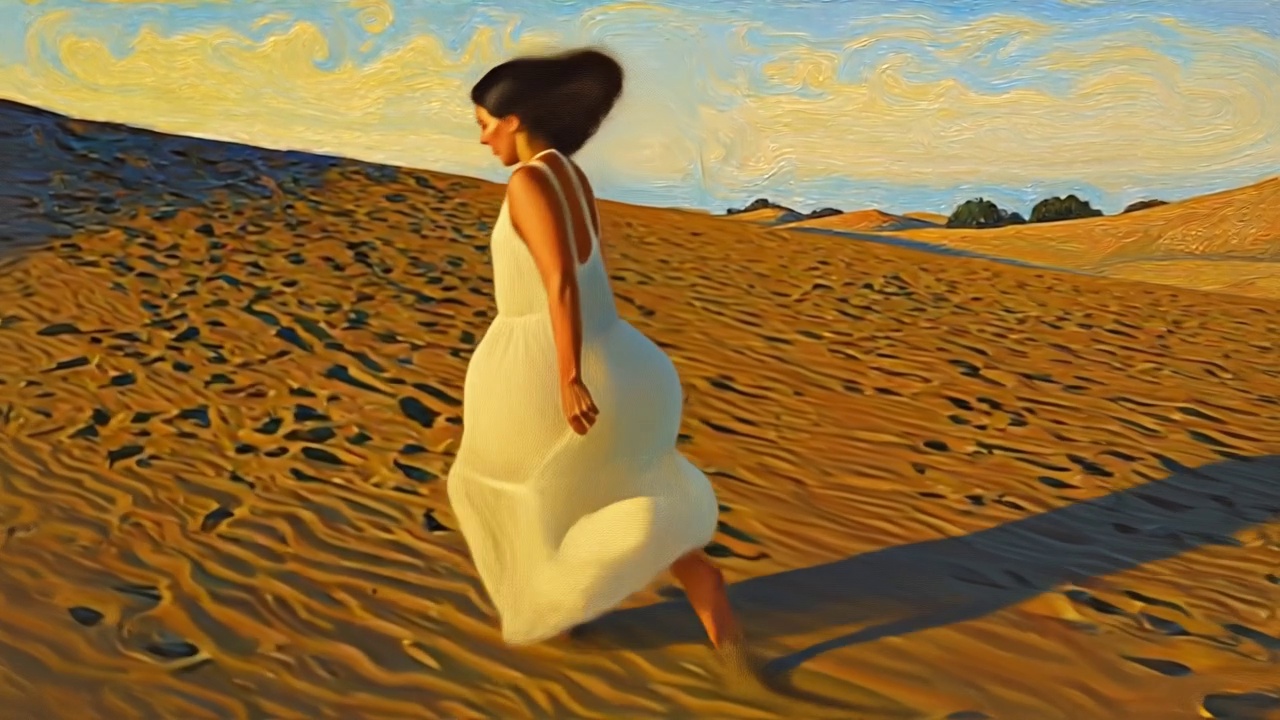} \\
        \includegraphics[width=0.33\linewidth]{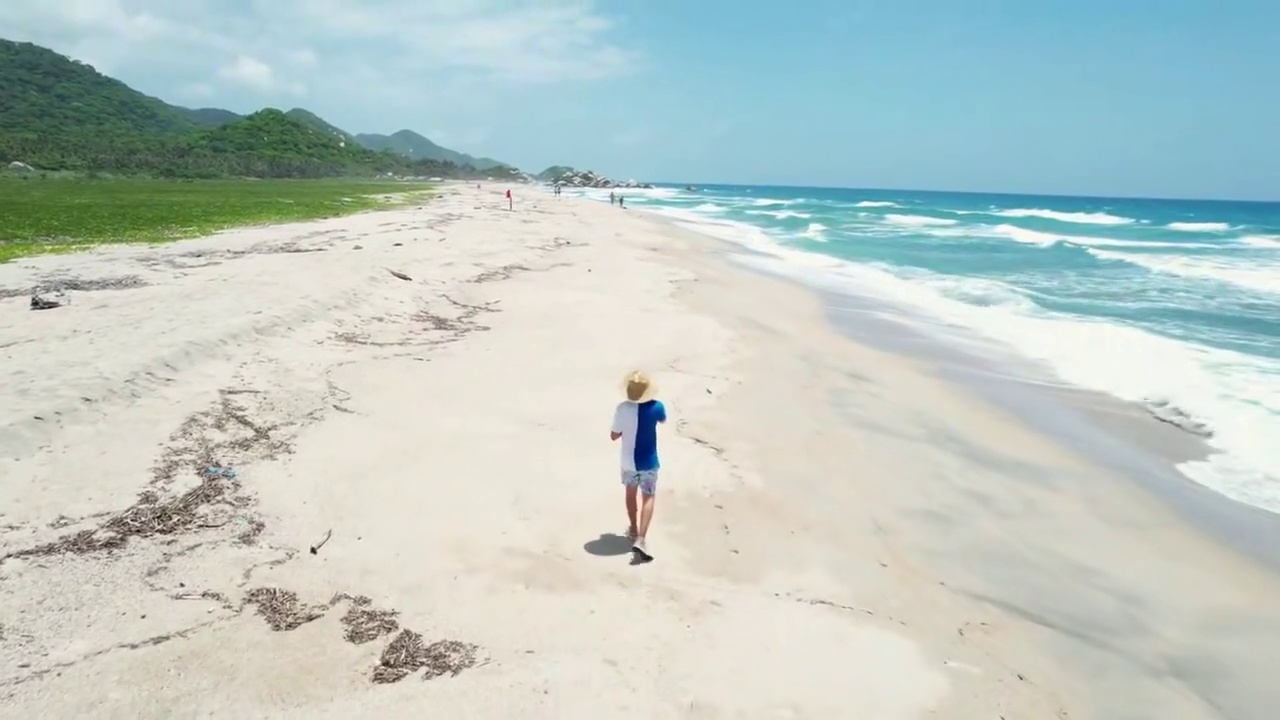} &
        \includegraphics[width=0.33\linewidth]{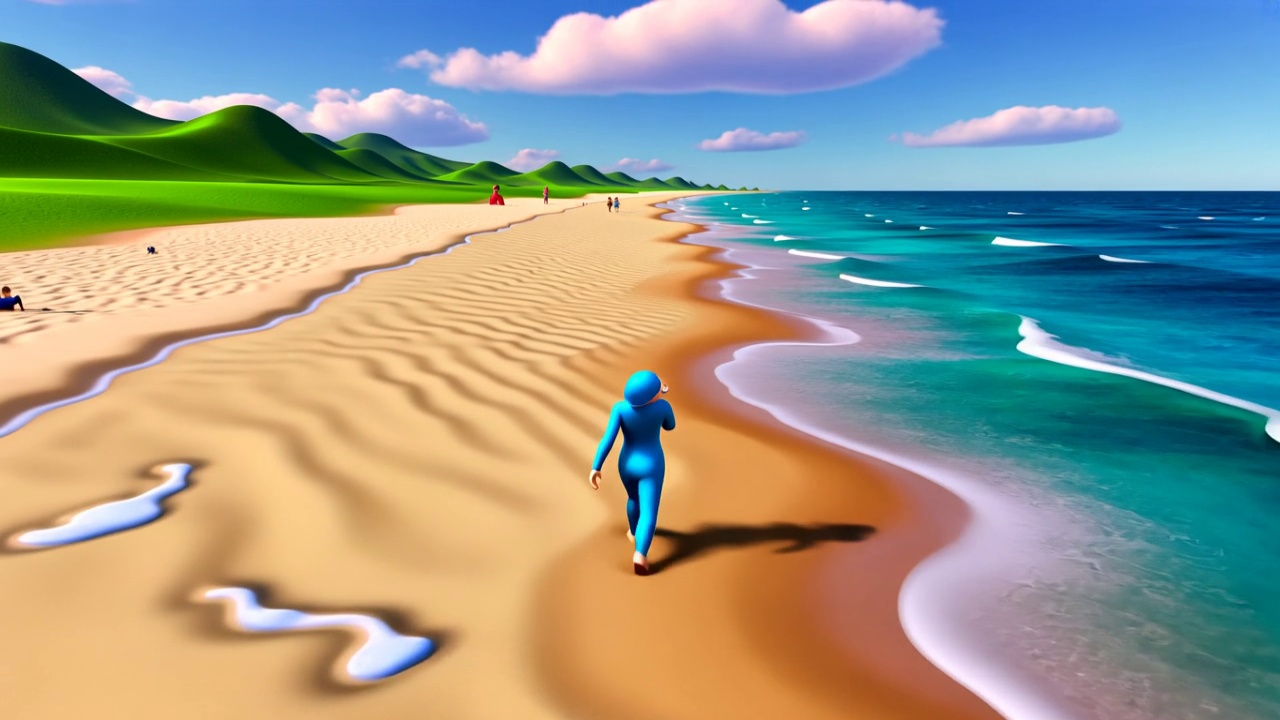} &
        \includegraphics[width=0.33\linewidth]{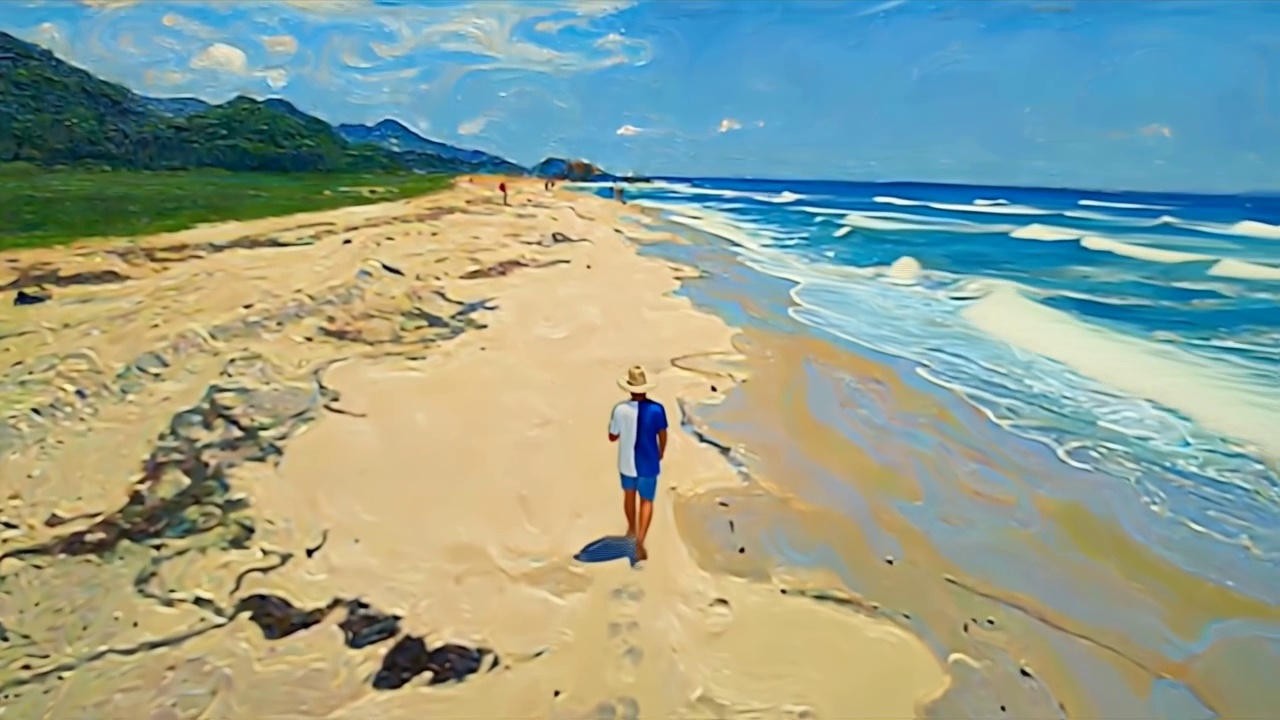} \\
        \includegraphics[width=0.33\linewidth]{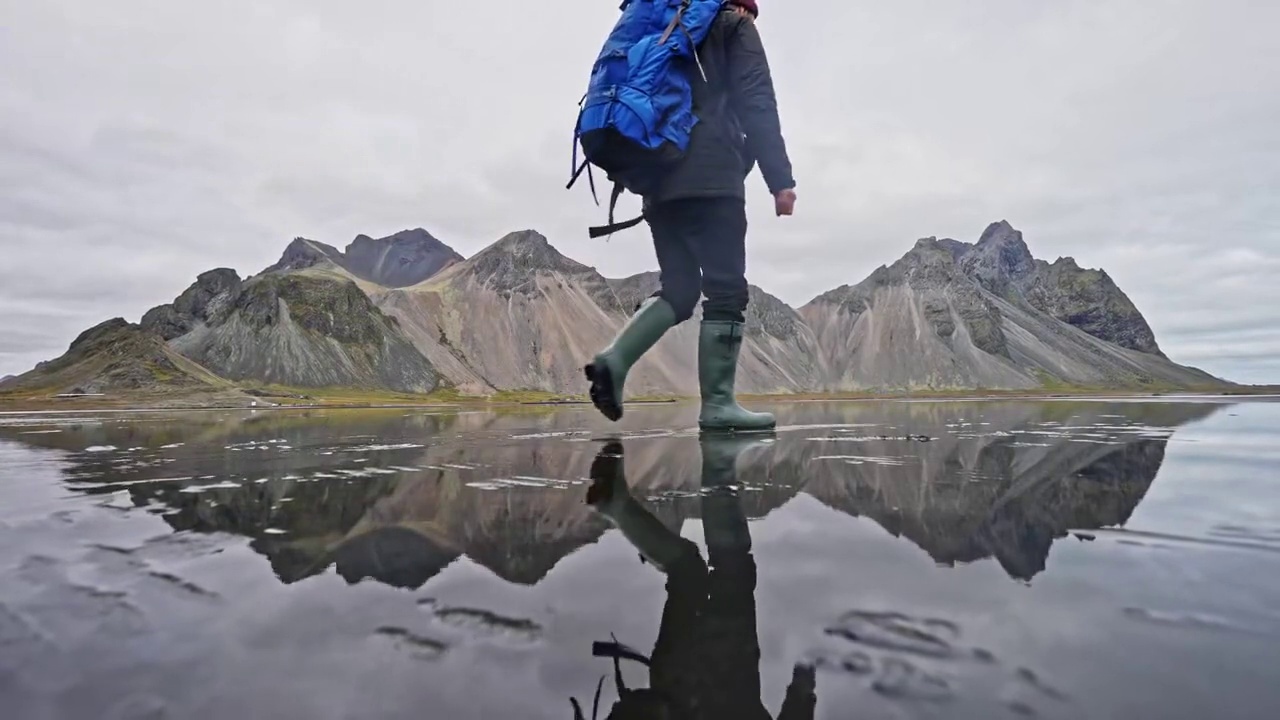} &
        \includegraphics[width=0.33\linewidth]{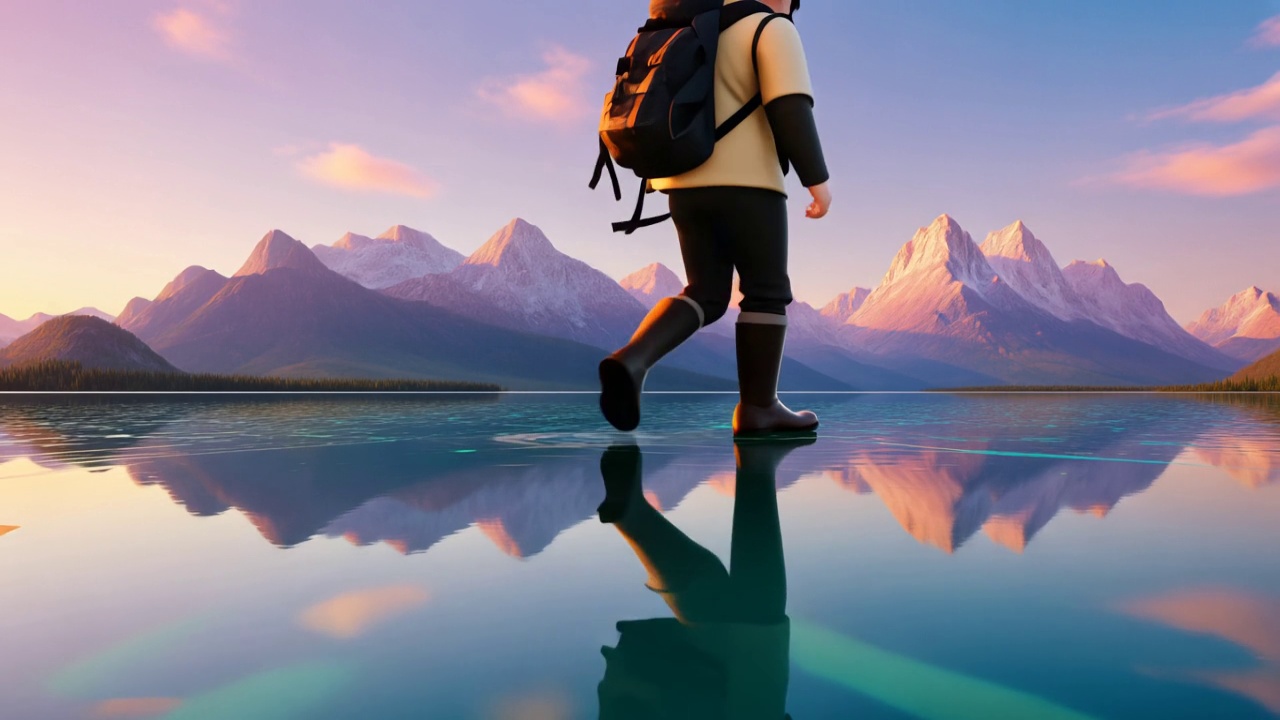} &
        \includegraphics[width=0.33\linewidth]{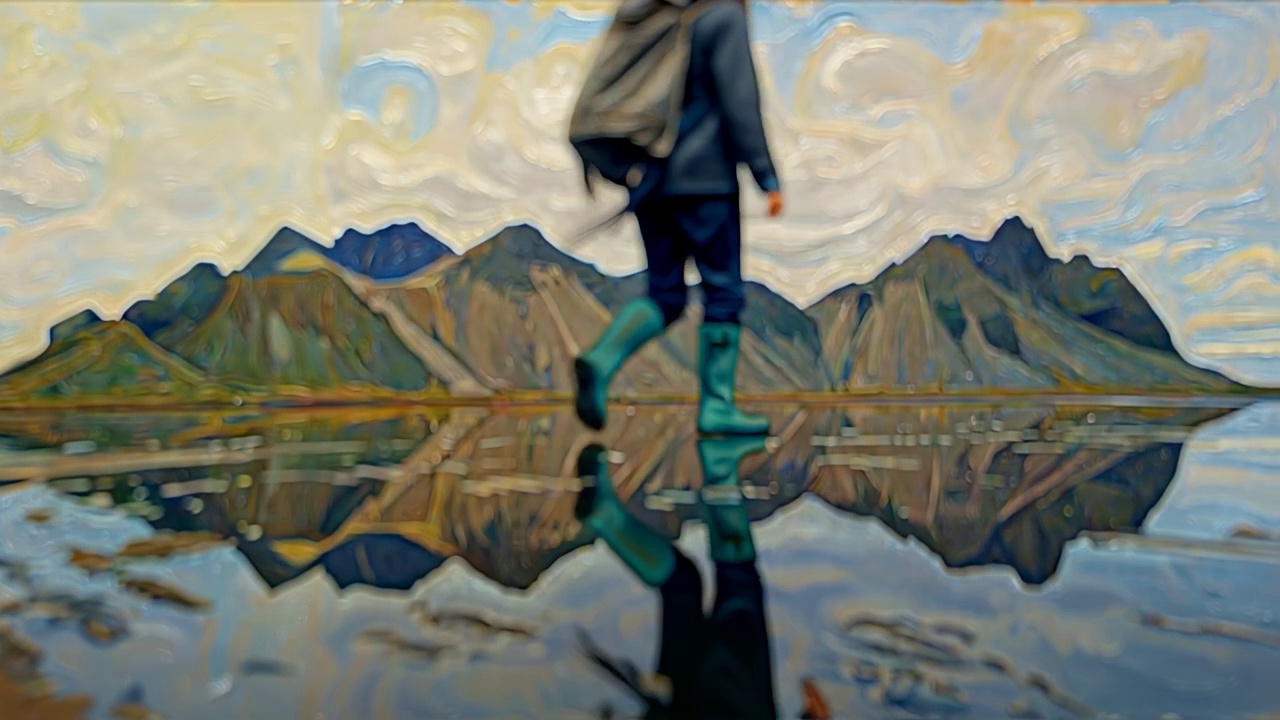} \\
        \includegraphics[width=0.33\linewidth]{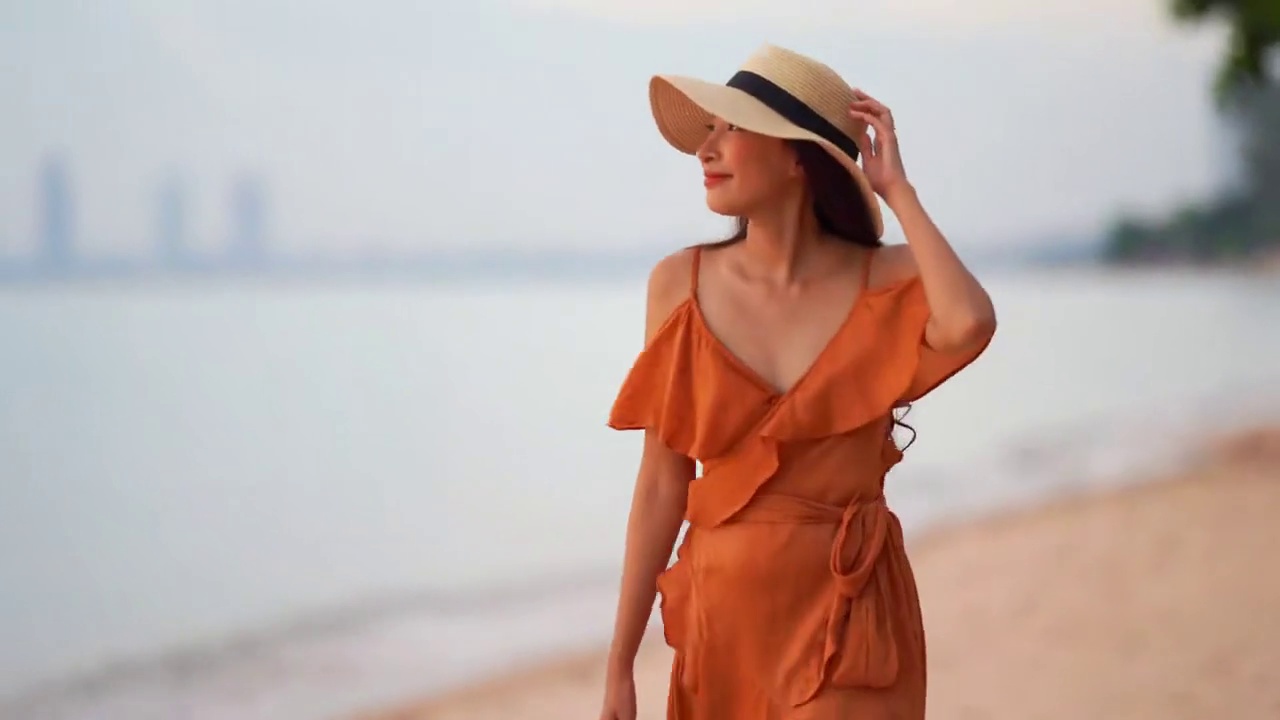} &
        \includegraphics[width=0.33\linewidth]{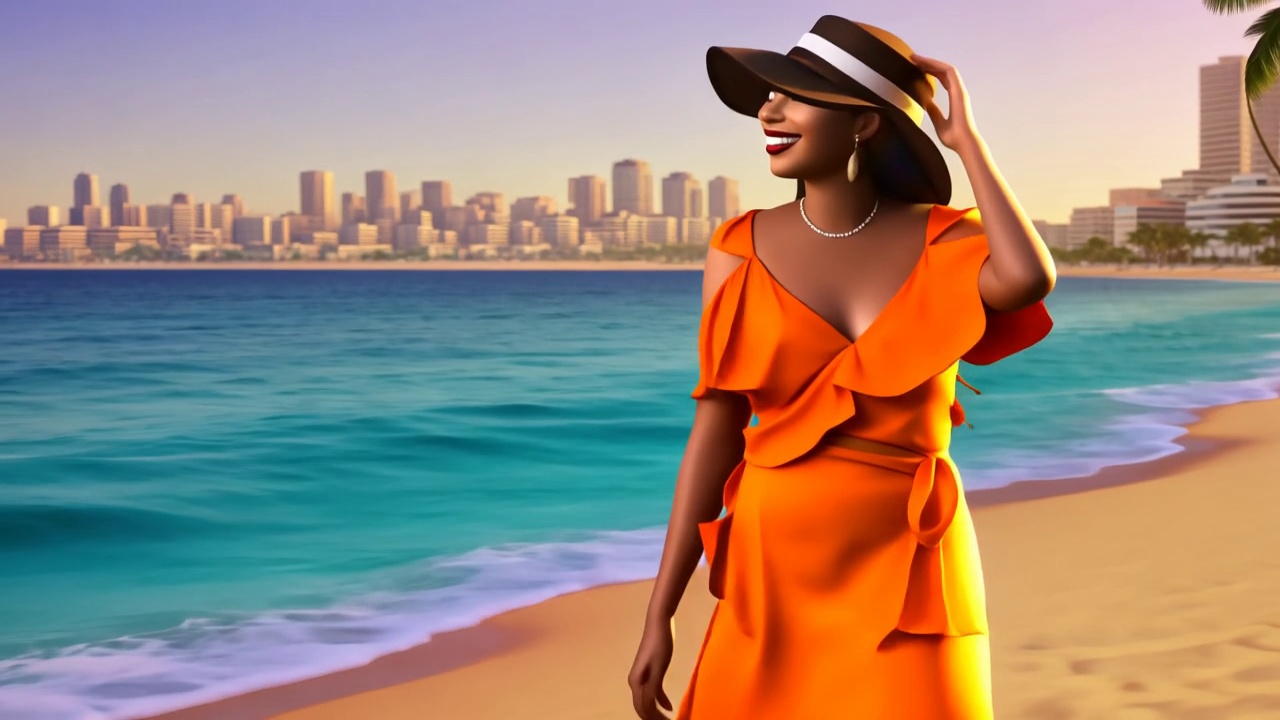} &
        \includegraphics[width=0.33\linewidth]{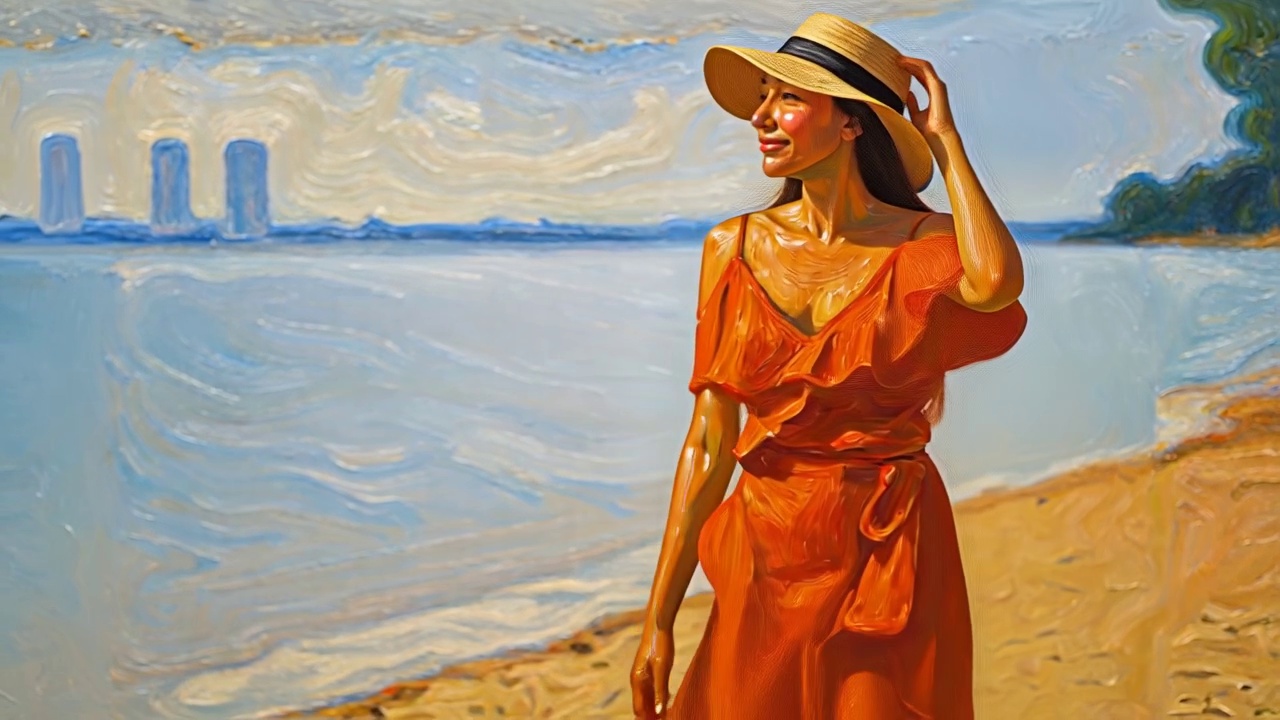} \\
    \end{tabular}
    \vspace{2pt}\textbf{(a)}
  \end{minipage}%
}%
\setlength{\Aht}{\ht\Abox}

\begin{minipage}[t]{0.48\textwidth}
  \centering
  \usebox{\Abox}
  \phantomsubcaption\label{fig:video_heatmap_comparison:a}
\end{minipage}
\hfill
\begin{minipage}[t]{0.48\textwidth}
  \centering
  \makebox[\linewidth][c]{%
    \resizebox{!}{\Aht}{%
      \begin{minipage}{0.99\linewidth}
        \centering

        \begin{tabular}{c c c c}
            \textbf{X-CLIP} & \textbf{SlowFast} &
            \textbf{VideoMoCo} & \textbf{MaCLR}\\
            \includegraphics[width=0.25\linewidth]{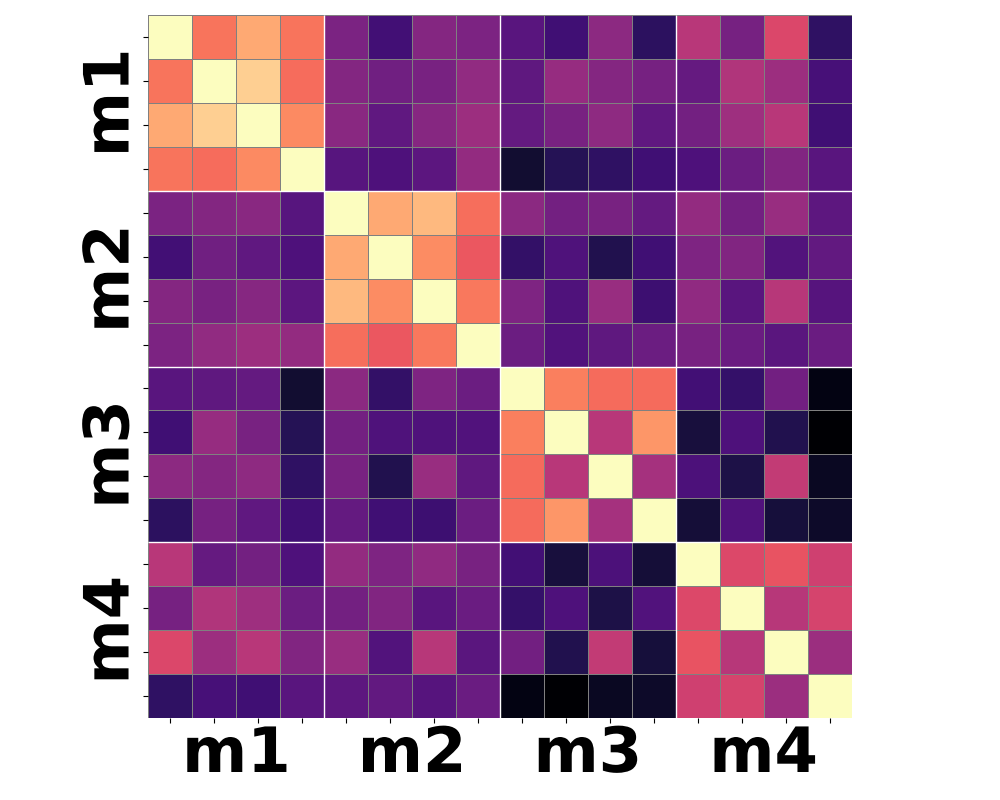} &
            \includegraphics[width=0.25\linewidth]{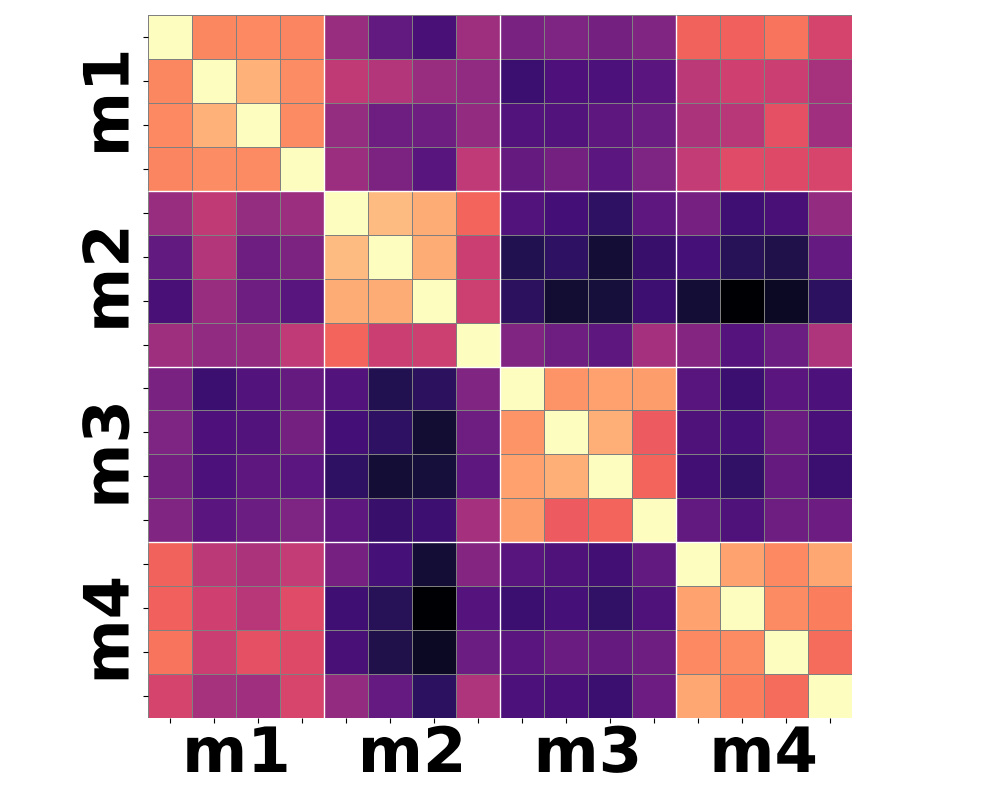} &
            \includegraphics[width=0.25\linewidth]{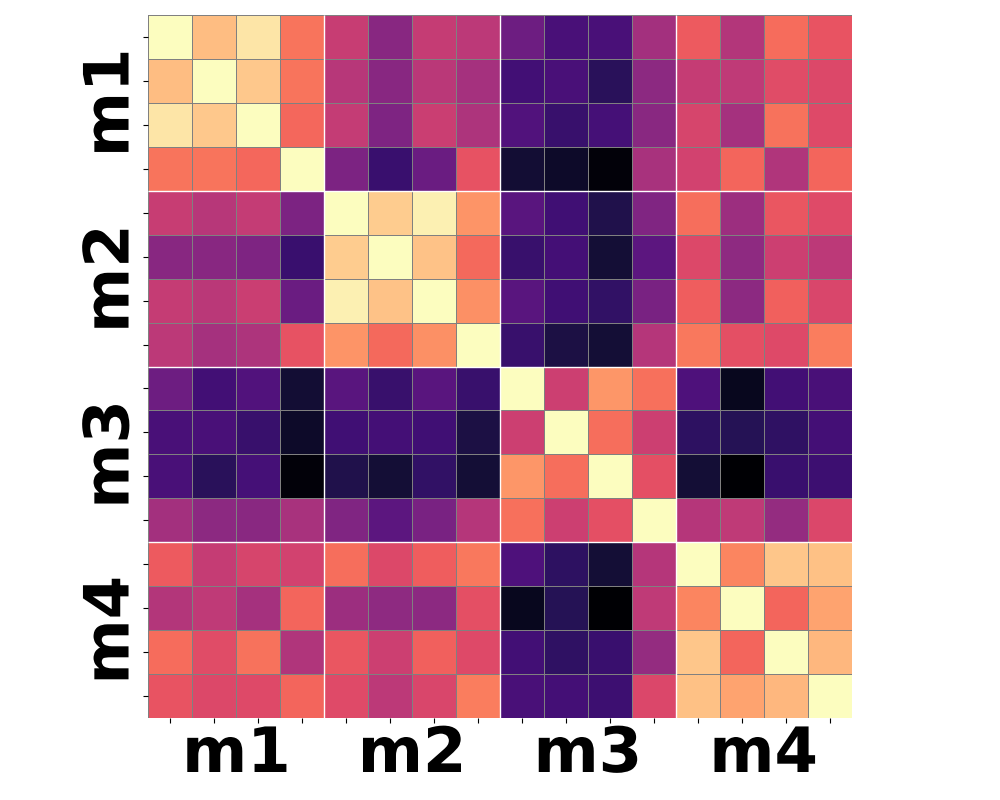} &
            \includegraphics[width=0.25\linewidth]{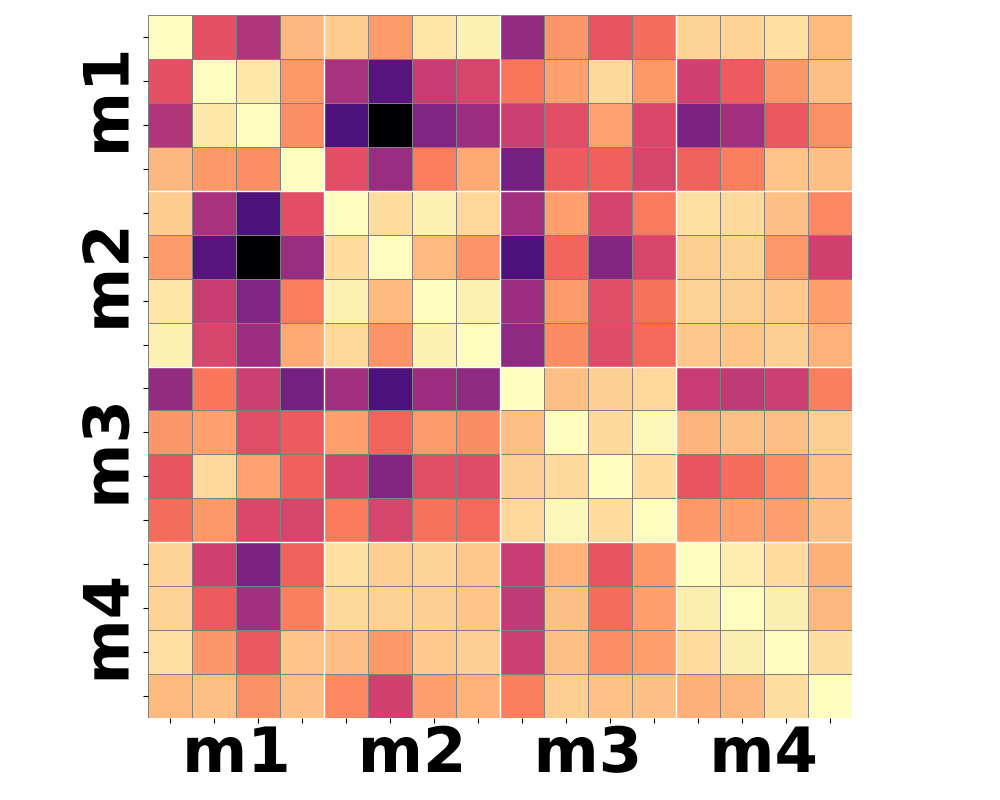} \\
        \end{tabular}
        \vspace{2pt}\textbf{(b)}
        \phantomsubcaption\label{fig:video_heatmap_comparison:b}

        \vspace{0.9em}

        \begin{tabular}{c c c c}
            & \textbf{DINOv2} & \textbf{VideoPrism} & \textbf{VideoMAE} \\
            \raisebox{7.5ex}{\rotatebox[origin=c]{90}{\textbf{Global Mean}}} &
            \includegraphics[width=0.3\linewidth]{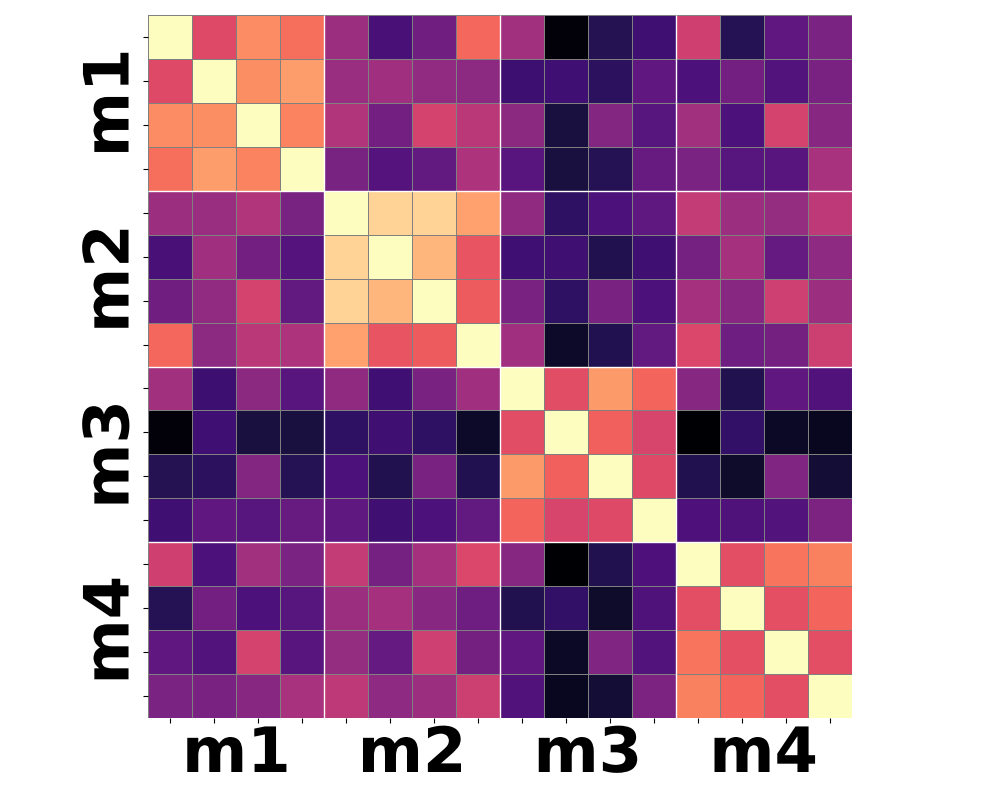} &
            \includegraphics[width=0.3\linewidth]{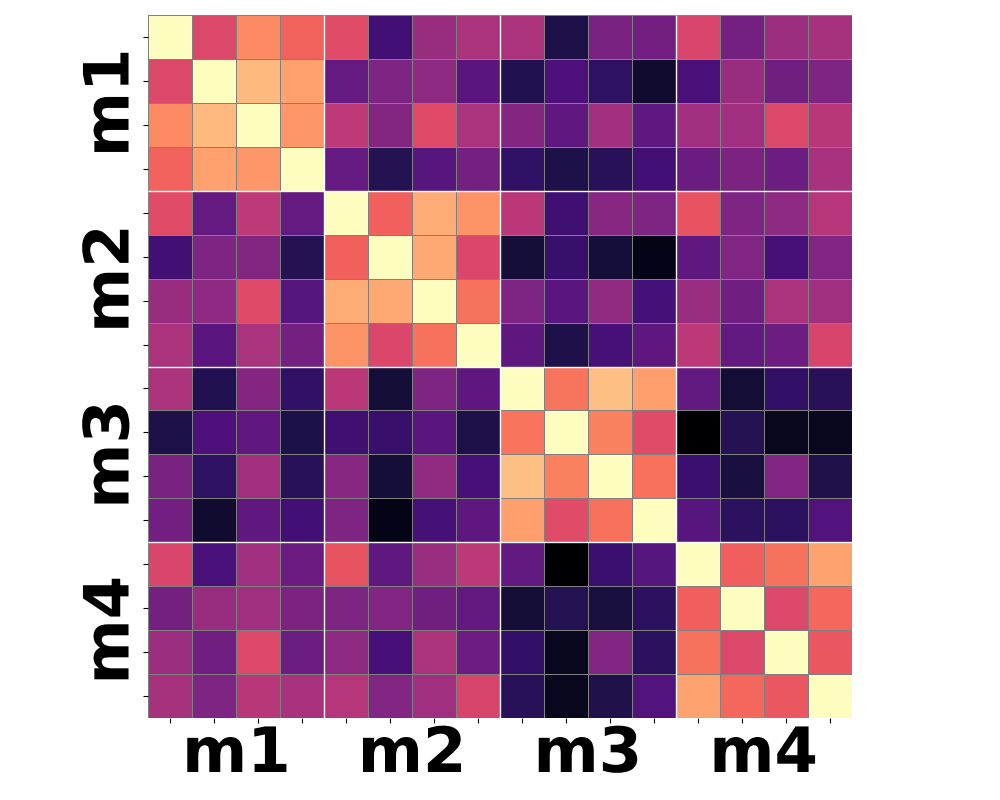} &
            \includegraphics[width=0.3\linewidth]{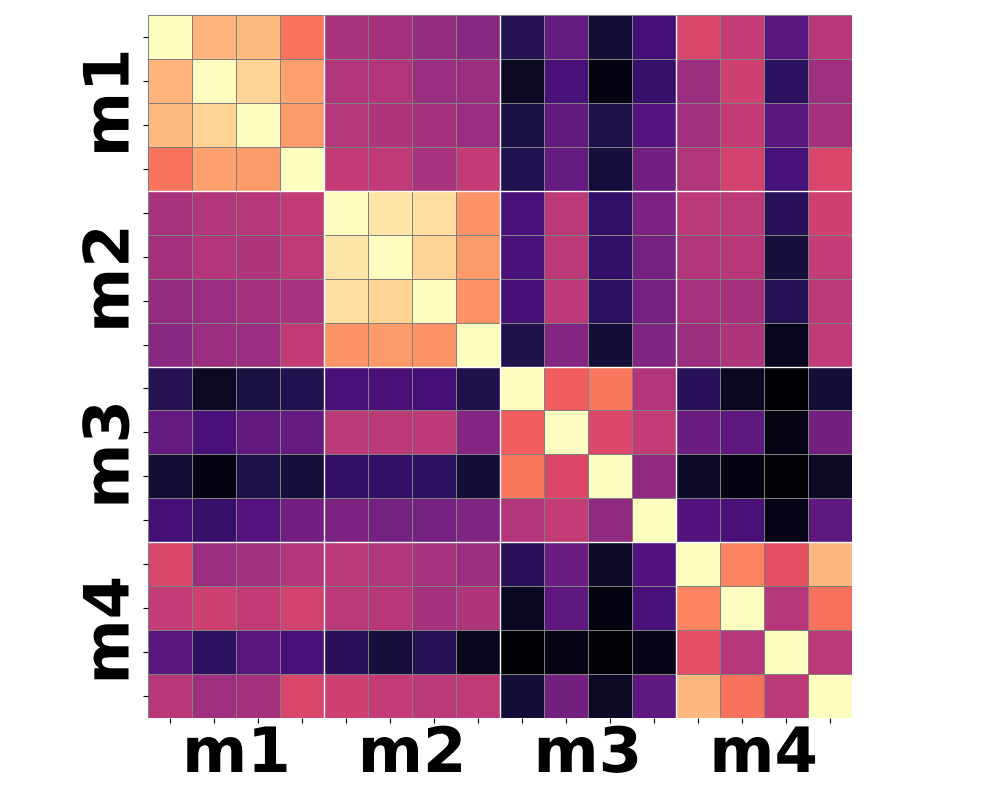} \\
            \raisebox{7.5ex}{\rotatebox[origin=c]{90}{\textbf{3rd Moment}}} &
            \includegraphics[width=0.3\linewidth]{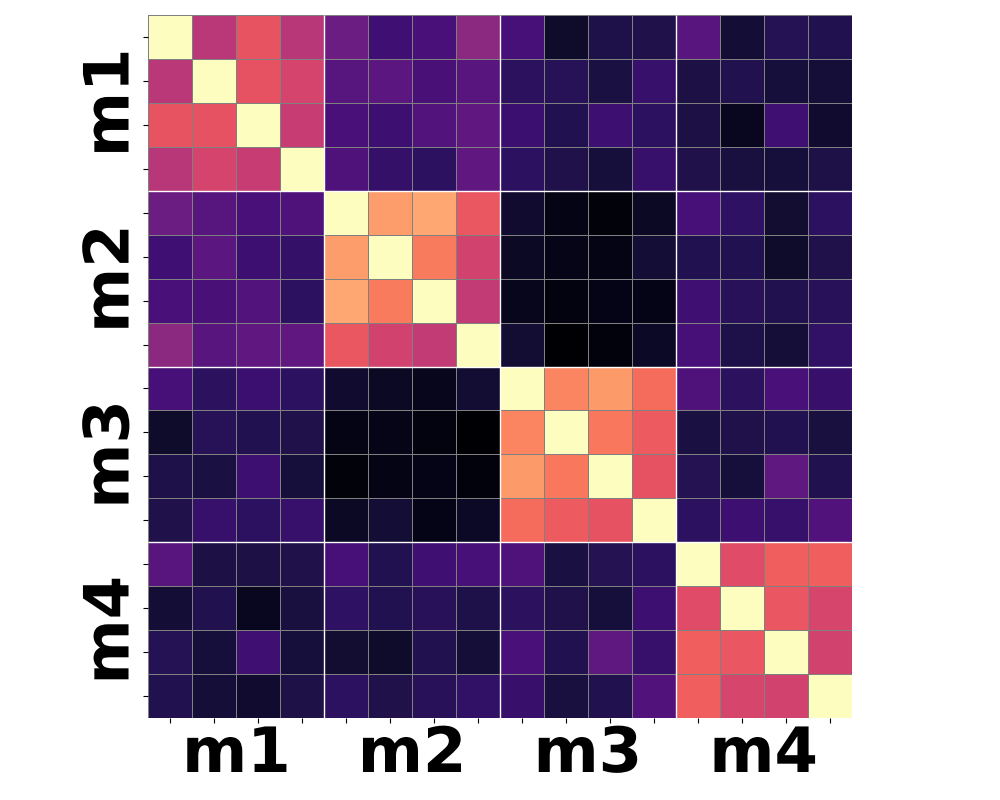} &
            \includegraphics[width=0.3\linewidth]{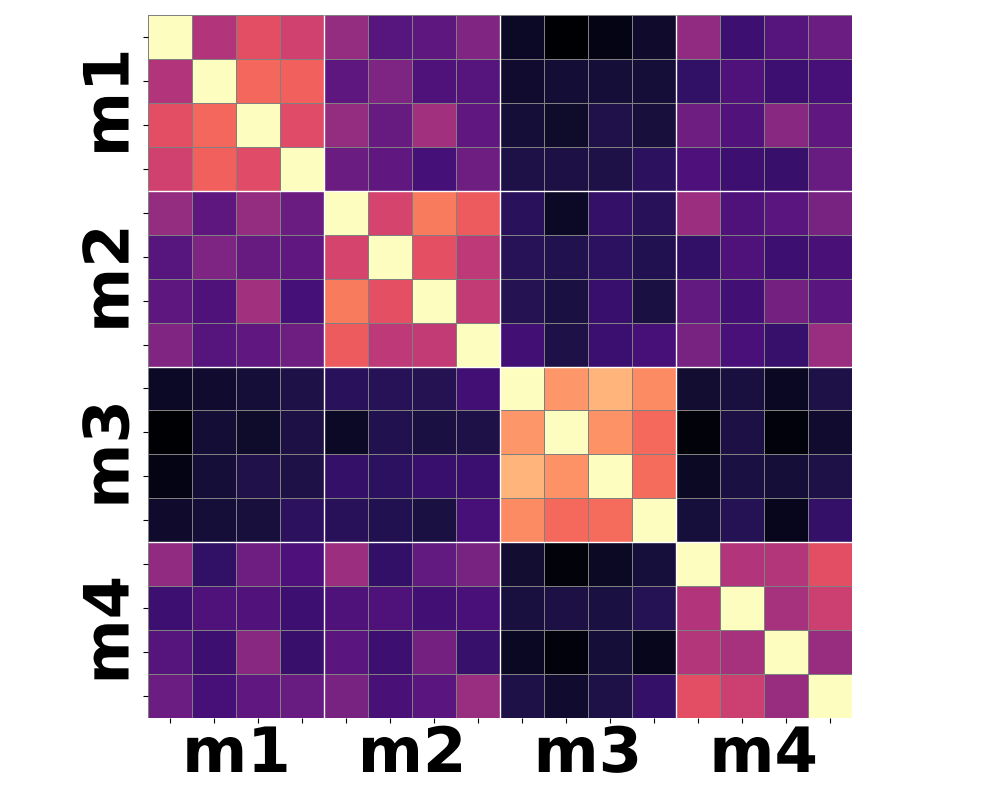} &
            \includegraphics[width=0.3\linewidth]{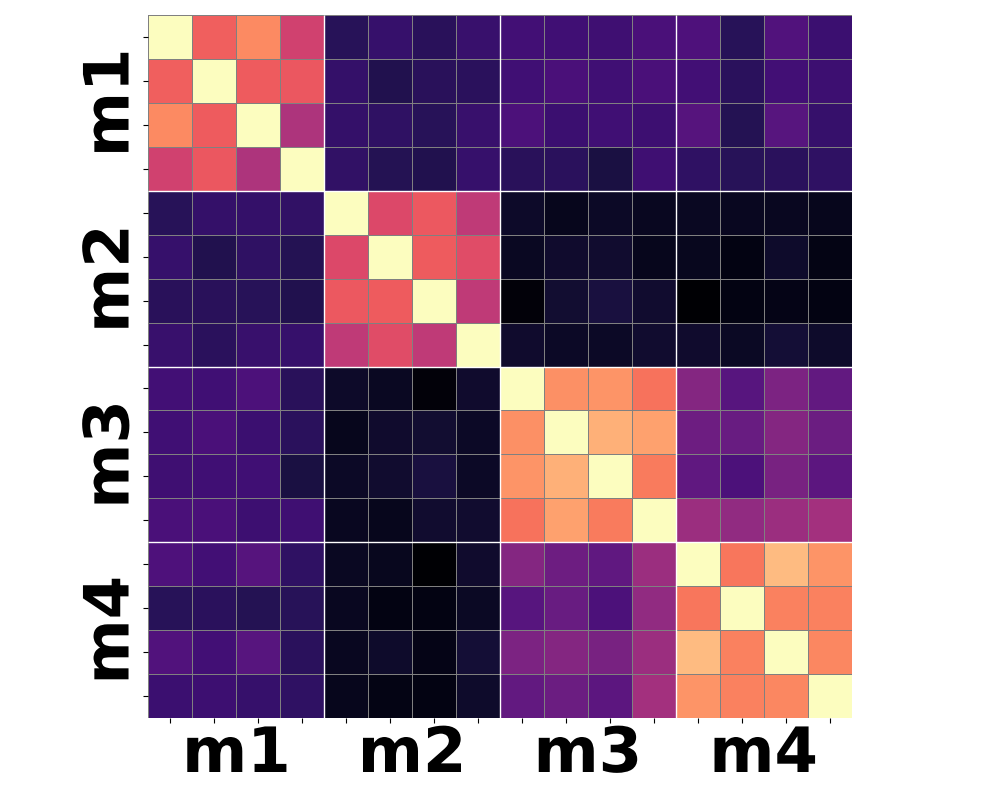} \\
        \end{tabular}
        \vspace{2pt} \\ \textbf{(c)}
        \phantomsubcaption\label{fig:video_heatmap_comparison:c}

      \end{minipage}
    }
  }
\end{minipage}
}
\vspace{-0.1cm}
\caption{\textbf{Motion-focused similarity with moment statistics.}
(a) Appearance-altered edits preserve the same underlying motion for each motion group $m_{i}$, while changing visual style.
(b) Baseline embeddings yield similarity heatmaps that are sensitive to appearance rather than motion.
(c) Our moment-based embedding (using the first three moments over patch features) produces clearer motion-consistent clusters (corresponding to shared motion 
$m_{i}$) than global mean pooling. Brighter cells indicate higher cosine similarity.
}
\vspace{-0.3cm}
\label{fig:video_heatmap_comparison}
\end{figure*}

To better evaluate motion similarity, we introduce the \textbf{\dataset} benchmarks, where similarity is defined through relative comparisons rather than action classification. Specifically, each \dataset{} benchmark includes both intra-class pairs, where videos share the same coarse motion type but differ in finer details, and inter-class pairs that differ in their motion altogether (e.g., running vs. jumping). This composition enables evaluation of both fine-grained similarity and broader action-type discrimination. 

Each benchmark serves a distinct purpose.
\textbf{\datasetS{}} serves as a controlled diagnostic benchmark, constructed to isolate and test the failure modes of existing representations. Its design allows us to systematically analyze how non-motion factors, such as appearance or viewpoint, influence similarity results while the underlying motion is held constant.
\textbf{\datasetR{}} complements this by testing generalization to real-world in-the-wild scenarios. It is manually curated and therefore smaller in scale. Its purpose is not to isolate specific factors, but to evaluate a model's alignment with human perception on realistic videos where motion is perceptually similar but not identical, and confounding factors like appearance and execution vary naturally. Both benchmarks will be made fully available and open source. In the following, we describe each benchmark in detail.

\begin{figure*}[t]
\begin{center}
  \centering
    \includegraphics[width=\textwidth]{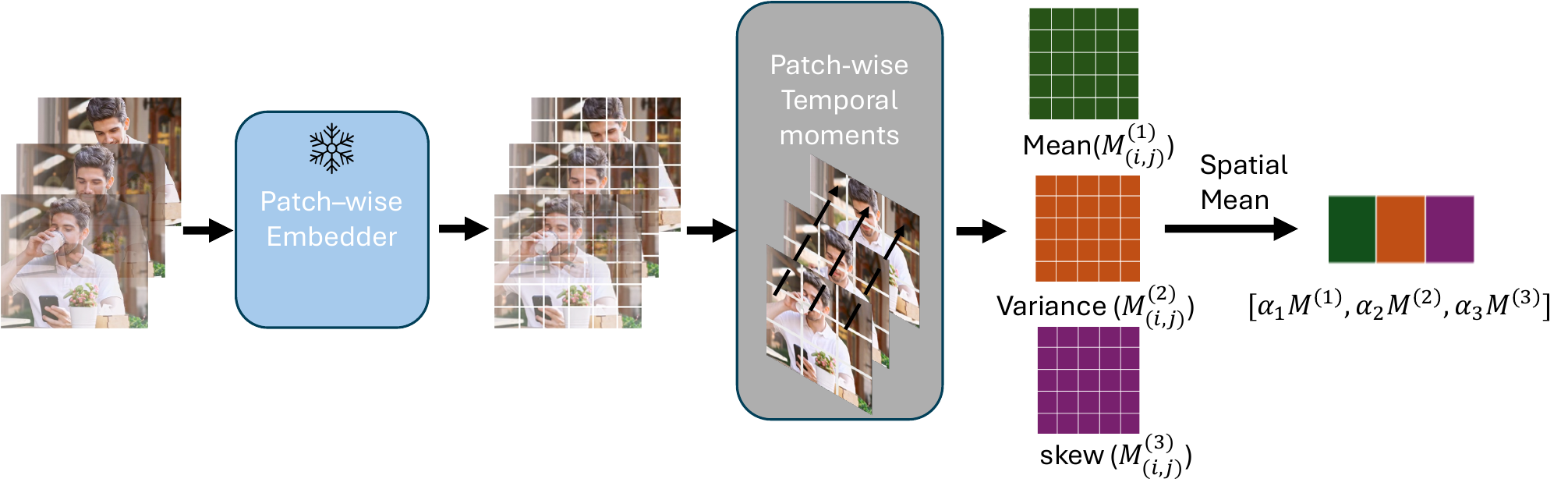}
  \vspace{-0.6cm}
  \captionof{figure}{\textbf{SemanticMoments pipeline.}
Patch-wise features are extracted per frame using a pretrained embedder (e.g., DINO) and summarized over time using the first three temporal moments (mean, variance, and skewness). Spatial aggregation yields one descriptor per moment, which are combined into a global motion-centric video embedding.}
  \label{fig:method_scheme}
\vspace{-0.5cm}
\end{center}
\end{figure*}

\subsection{\datasetS}
\label{synthetic_dataset}

\noindent\textbf{Benchmark Structure.}
The benchmark contains 250 triplets (750 videos in total), each composed of a reference video, a positive that shares the same motion, and a hard negative with similar appearance but different motion.
Triplets are organized into five categories, each defining how the positive video differs from the reference while preserving motion, with 50 instances per category:
(1)~\emph{Static Object}, where non-moving elements in the scene are added, removed, or replaced; 
(2)~\emph{Dynamic Object}, where the main moving subject is replaced by another entity performing the same motion; 
(3)~\emph{Dynamic Appearance}, where visual attributes such as clothing, tattoos, or accessories are modified on the moving subject; 
(4)~\emph{Scene Style}, where the rendering style of the scene is changed (e.g., realistic, painting, or sketch); and 
(5)~\emph{View}, where the camera viewpoint is altered. Visual examples of these categories are shown in Fig.~\ref{fig:sync_motions_dataset}. 
Each category isolates a different type of variation, with positives preserving motion dynamics and negatives sharing appearance but differing in motion.

\noindent\textbf{Generation Pipeline.}
For each triplet, we first generate four textual prompts using GPT-4.1~\cite{achiam2023gpt}:
(1) a base prompt describing the scene,
(2) a prompt defining the modification in the scene according to the category,
(3) a video prompt specifying the intended motion, and
(4) a negative motion prompt describing a distinct motion for the same subject.

Prompts (1) and (2) are used to synthesize reference and positive images with Gemini2.5-Flash~\cite{comanici2025gemini}, which are concatenated and passed to WAN 2.2 (Image-to-Video)~\cite{wan2025} to generate temporally synchronized videos sharing identical motion using a single, shared prompt (3). We follow a synchronization strategy, conceptually similar to IC-LoRA~\cite{huang2024context}, which ensures WAN 2.2 applies an identical temporal transformation to both (semantically-aligned) start frames. This results in two videos that are temporally synchronized and share the exact same motion dynamics while differing only in the specified visual factors.

Finally, a hard negative video is produced from the same base image using prompt (4), yielding identical appearance but different motion.
This pipeline ensures strong temporal alignment while enabling systematic control over appearance, subject, and viewpoint.
All videos are 5 seconds long, sampled at 16 fps with a spatial resolution of 512×512. 

For evaluation, all other videos in the benchmark are treated as additional negatives during retrieval, providing a diverse pool and increasing the difficulty of motion discrimination. This setup provides a clean and controlled basis for evaluating motion representations.

\subsection{\datasetR}

The benchmark contains 40 examples, each centered on a reference video paired with a positive and a negative counterpart. The positive video depicts the same underlying motion despite differences in appearance or context, while the negative shares a similar appearance but differs in motion. Negative pairs are obtained by sampling different short clips with different motion from the same source video, ensuring comparable visual context. Positive candidates are retrieved from Pexels~\cite{pexels_website} using text-based motion descriptions and ranked through crowd-sourced annotation, where annotators judged which clips exhibited the most similar motion to the reference, regardless of appearance or scene differences (see supplementary for further details). 
This process grounds similarity in human perception of motion rather than in visual or categorical cues, yielding a benchmark that reflects naturally occurring motion variability beyond the controlled design of \datasetS.
In addition to the hard negatives, we include 
randomly sampled videos 
from the Kinetics-400 test set as \emph{random negatives}. Together, these examples form a realistic benchmark for evaluating whether representations can generalize to motion similarity under natural variation. 

\section{Analysis}
\label{sec:analysis}
Prior work primarily optimizes for action recognition or video–text alignment, which under-specify temporal structure. In addition, self-supervised approaches such as masked autoencoding often under-represent motion and remain biased toward appearance. We therefore hypothesize that embeddings learned from these objectives are suboptimal for modeling motion similarity.

To probe this limitation, we construct a controlled set of four walking sequences and synthesize motion-preserving variants that vary only in style while maintaining identical dynamics, as can be seen in \cref{fig:video_heatmap_comparison:a}. These controlled variations isolate motion similarity as the sole factor of interest.

Ideally, a motion-sensitive representation should cluster each clip with its motion-equivalent variants while distinguishing distinct walking styles. We evaluate this by computing pairwise cosine-similarity matrices over video embeddings and visualizing them as heatmaps. As shown in~\cref{fig:video_heatmap_comparison:b}, existing methods show partial grouping but fail to consistently isolate shared motion across different styles.

We conclude that embeddings from self-supervised image and video encoders (e.g., DINOv2, VideoPrism) fail to consistently capture fine-grained motion similarity. Leveraging their semantic correspondence and summarizing temporal variation with higher-order moments (e.g., variance, skewness) yields motion-sensitive embeddings that correctly cluster motion-equivalent variants while distinguishing distinct walking style, as we demonstrate in ~\cref{fig:video_heatmap_comparison:c}.

\section{\algname{}}

Our analysis in Sec.~\ref{sec:analysis} reveals that current video encoders tend to focus on appearance and scene context. Their embeddings shift noticeably in feature space under style-only edits that preserve geometry and motion. Our key observation is that incorporating higher-order temporal moments result in features that better represent motion, as such embeddings yielding similarities that better reflect underlying semantic motion patterns.
Motivated by this, we define in Sec.~\ref{sec:M+} a general, moment-based representation space $\mathcal{M}+$ that encodes temporal statistics into structured video descriptors
We then instantiate it in Sec.~\ref{sec:alg} with \algname{}, a practical, training-free method that aggregates the first three temporal moments of pretrained features to form motion-centric representations (see Fig.~\ref{fig:method_scheme}). Finally, in Sec.~\ref{sec:experiments} we evaluate \algname{}, demonstrating strong performance on motion-based video retrieval.


\subsection{\texorpdfstring{$\mathbf{\mathcal{M}+}$}{M+}: A Parametric View of Temporal Statistics}
\label{sec:M+}
We define $\mathcal{M}+$, a \textit{moment space} over temporal visual features, extending conventional video representations from a single embedding to a structured set of temporal descriptors. Rather than collapsing temporal information into one pooled vector, we compute multiple statistical moments across time, where each moment captures a distinct temporal characteristic.

\paragraph{Patch-wise temporal moments.}
Given a Video $\mathcal{C}$ and feature extractor $\mathcal{F}$, feeding each frame $t\in\{1,\dots,T\}$ to the feature extractor produce $P$ patch features
$F_t\in\mathbb{R}^{P\times d}$. We denote the $d$-dimensional feature of patch $p$ by $f_{t,p}\in\mathbb{R}^d$.
We define the first temporal moment as the (non-central) mean:
\[
\mu^{(1)}_{p} = \mu_p = \frac{1}{T}\sum_{t=1}^T f_{t,p}.
\]

For higher-order orders $k>1$, we compute central temporal moments:
\[
\mu^{(k)}_{p} = \frac{1}{T}\sum_{t=1}^T (f_{t,p} - \mu_p)^k .
\]

Intuitively, $\mu^{(1)}_p$ encodes the average appearance of patch $p$ across time, 
$\mu^{(2)}_p$ captures the magnitude of temporal variation (motion energy), 
and $\mu^{(3)}_p$ reflects the directional asymmetry of change (motion polarity).

\paragraph{Spatial aggregation.}
So far, moments consists of significant spatial representation. To obtain a global motion representation per moment order, we aggregate the per-patch moments spatially:

\vspace{-0.4cm}
\[
M^{(k)} \;=\; \frac{1}{P}\sum_{p=1}^{P} \mu^{(k)}_{p} \;\in\; \mathbb{R}^{d}, \quad k=1,\dots,K.
\]
This yields one descriptor $M^{(k)}$ per statistical moment, each summarizing a distinct aspect of temporal variation.

\paragraph{Moment embedding.}
The general form of the \textbf{M+} video-level representation is obtained by a weighted concatenation of the different moment vectors, where $\alpha_k \in \mathbb{R}$ is the relative contribution of the $k$-th moment:

\vspace{-0.3cm}
\begin{equation*}
\phi_{\text{video}} \;=\; [\,\alpha_1 M^{(1)};\; \alpha_2 M^{(2)};\; \dots;\; \alpha_K M^{(K)}\,] \in \mathbb{R}^{Kd},
\end{equation*}

\subsection{Design choices}
\label{sec:alg}

We now present our video representation, designed to excel in motion-focused video similarity.
We operationalize M+ using pretrained semantic backbones: DINOv2, VideoMAE, and VideoPrism. 
We use the first three temporal moments ($k=1,2,3$): 
$M^{(1)}$ corresponds to average pooling, while $M^{(2)}$ and $M^{(3)}$ capture the magnitude and polarity of temporal change.
The moment vectors are concatenated using weights $\alpha_k$, yielding a final embedding 
$\phi_{\text{video}} \in \mathbb{R}^{3d}$.

For all experiments, we sample $T=32$ frames uniformly per video at the backbone’s native resolution 
and compute patch features from the final encoder layer.
Unless otherwise stated, we use $\alpha_1 = 1$, $\alpha_2 = 8$, and $\alpha_3 = 4$.
The entire process is training-free and introduces minimal additional computational cost, making it scalable to large-scale video collections.

\section{Experiments}
\label{sec:experiments}
We evaluate the effectiveness of our method on both synthetic and real-world motion-similarity benchmarks introduced in Sec.~\ref{sec:datasets}. We begin by outlining the baselines and evaluation protocol, followed by quantitative results on SimMotion-Synthetic (Sec.~\ref{section:quantitative_SimMotion_Synthetic}) and SimMotion-Real (Sec.~\ref{section:quantitative_SimMotion_Real}). Finally, we present ablation studies (Sec.~\ref{section:ablation_studies}) and discuss limitations (Sec.~\ref{section:limitations}).
Videos and additional results are provided in the supplementary materials.

\paragraph{\textbf{Baselines.}} 
We compare our method against a range of established approaches in video representation learning. Multimodal retrieval models, including CLIP4Clip~\cite{luo2022clip4clip} and X-CLIP~\cite{ma2022x}, leverage large-scale image–text pretraining for video retrieval. Optical-flow–based models such as I3D~\cite{carreira2017quo}, CoCLR~\cite{han2020self}, and MaCLR~\cite{xiao2022maclr} explicitly incorporate motion via flow. I3D and CoCLR use two-stream RGB–flow architectures, while MaCLR employs flow supervision only during training to guide motion-aware features. RGB-based supervised architectures, including SlowFast~\cite{feichtenhofer2019slowfast} and TimeSformer~\cite{gberta_2021_ICML}, are trained on large-scale action recognition datasets (e.g., Kinetics~\cite{kay2017kinetics}) to learn spatiotemporal representations directly from frames. Self-supervised transformer encoders such as VideoMAE~\cite{tong2022videomae} and VideoPrism~\cite{zhao2024videoprism} are pretrained with masked reconstruction objectives, and DINOv2~\cite{oquab2023dinov2} serves as a strong image-only self-supervised baseline. We use the publicly available implementations for all methods. Together, these baselines cover most of video representation techniques, enabling a comprehensive evaluation of our moment-based approach.

\begin{figure*}
    \centering
    \setlength{\tabcolsep}{0pt}
    {\scriptsize
    \begin{tabular}{ccc c ccc}
        \multicolumn{3}{c}{Reference video} && \multicolumn{3}{c}{Reference video} \\
        \includegraphics[width=0.165\textwidth, height=0.087\textwidth]{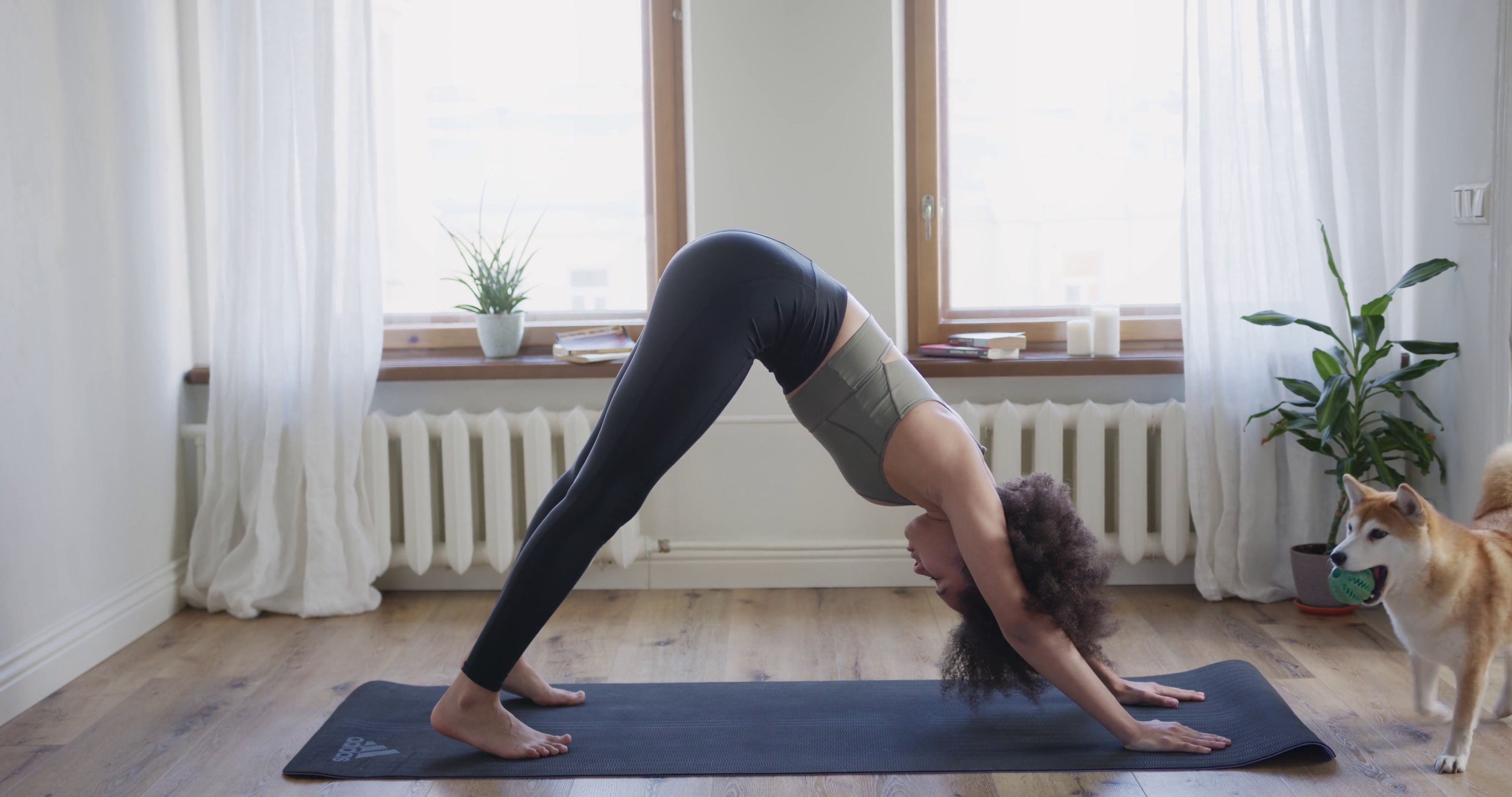} &
        \includegraphics[width=0.165\textwidth, height=0.087\textwidth]{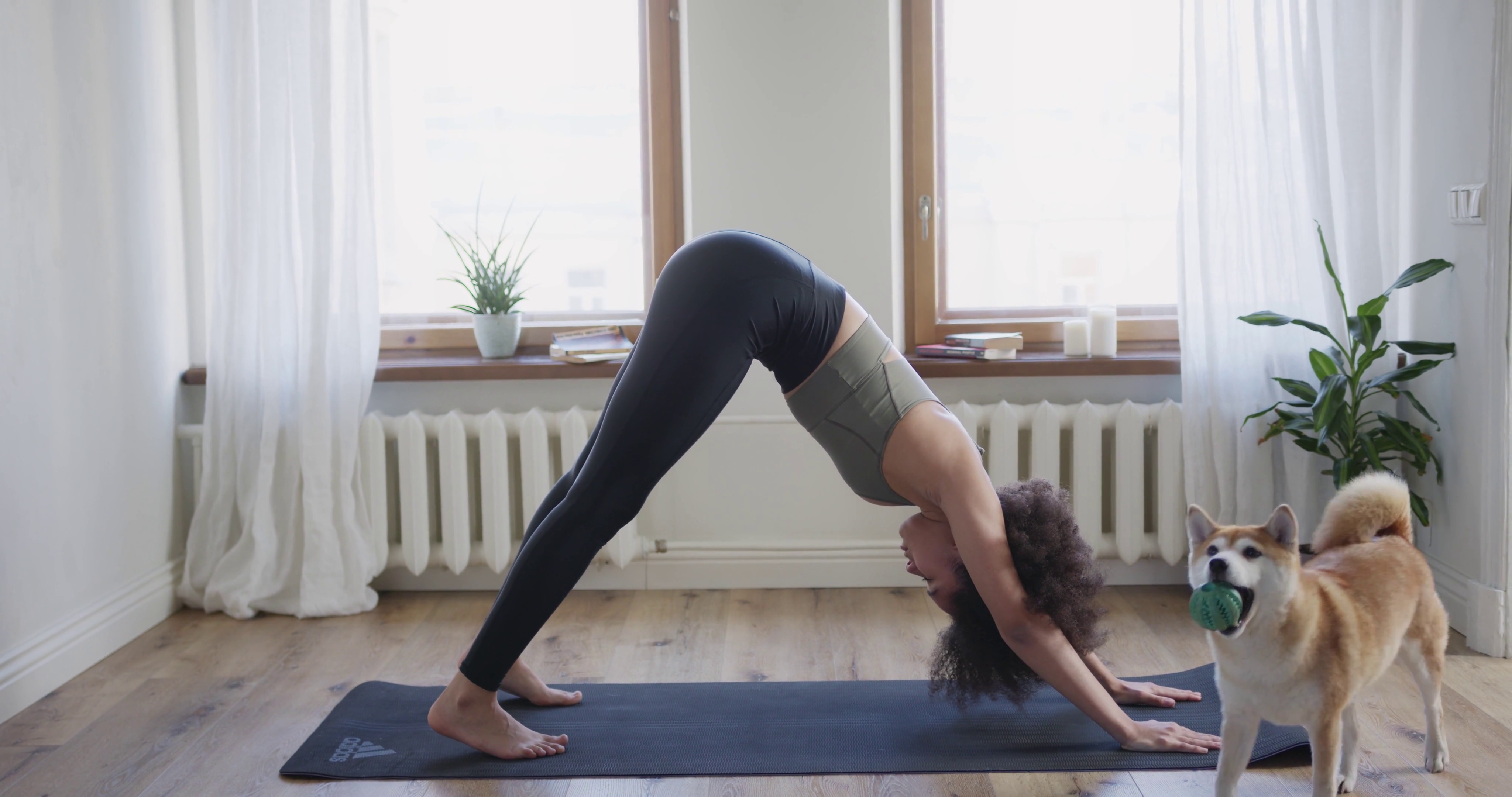} &
        \includegraphics[width=0.165\textwidth, height=0.087\textwidth]{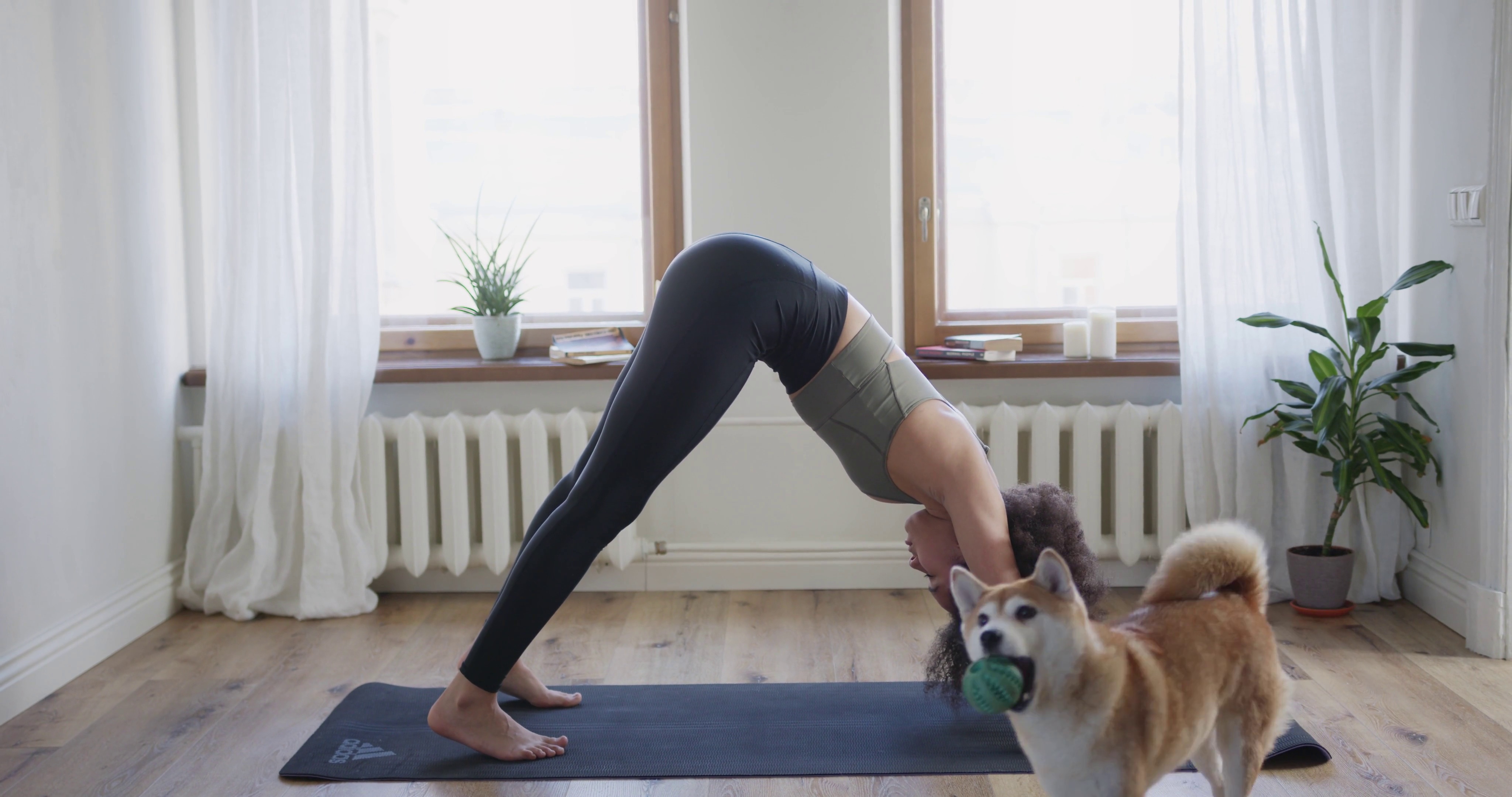} &
        {} &
        \includegraphics[width=0.165\textwidth, height=0.087\textwidth]{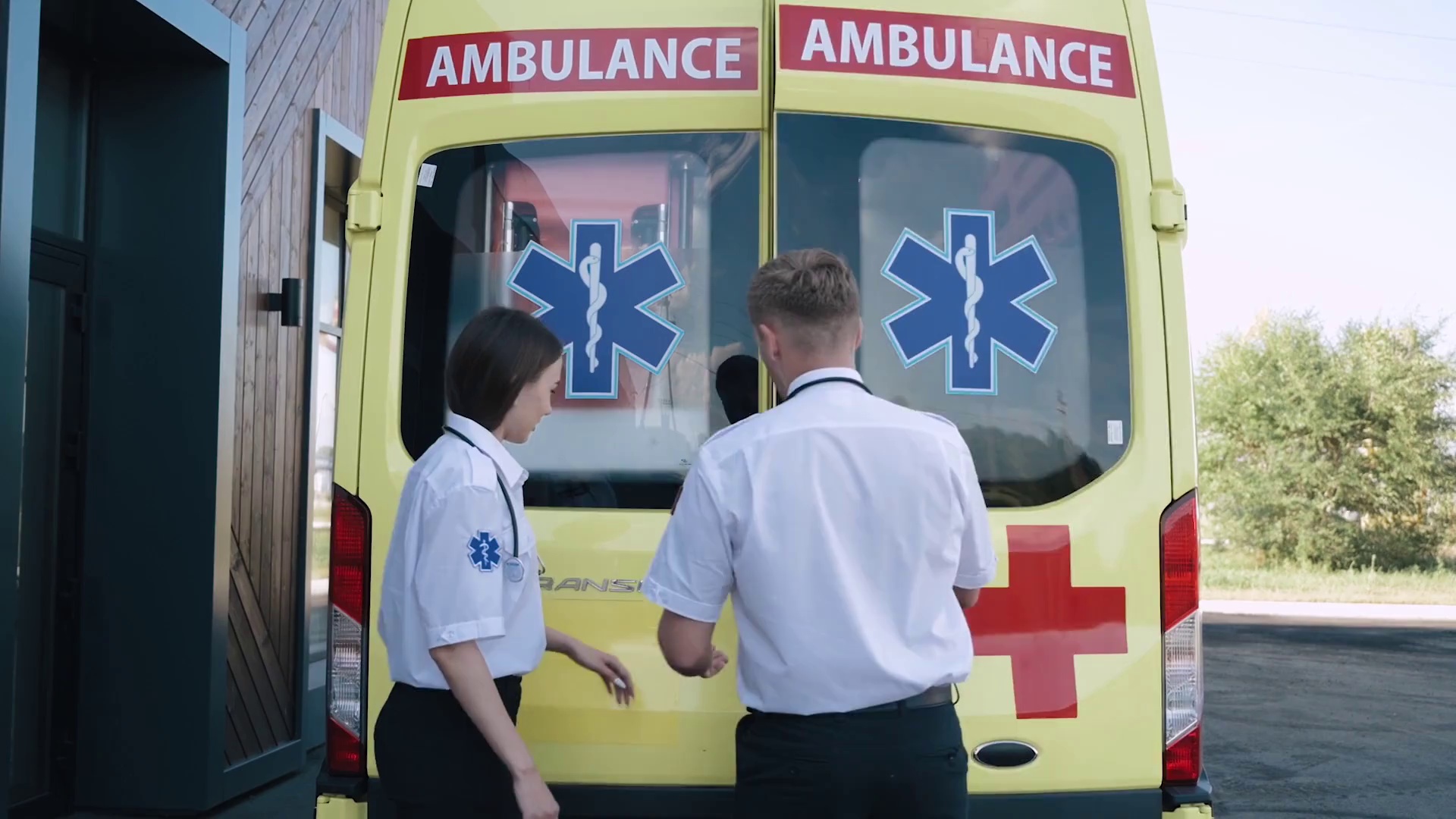} &
        \includegraphics[width=0.165\textwidth, height=0.087\textwidth]{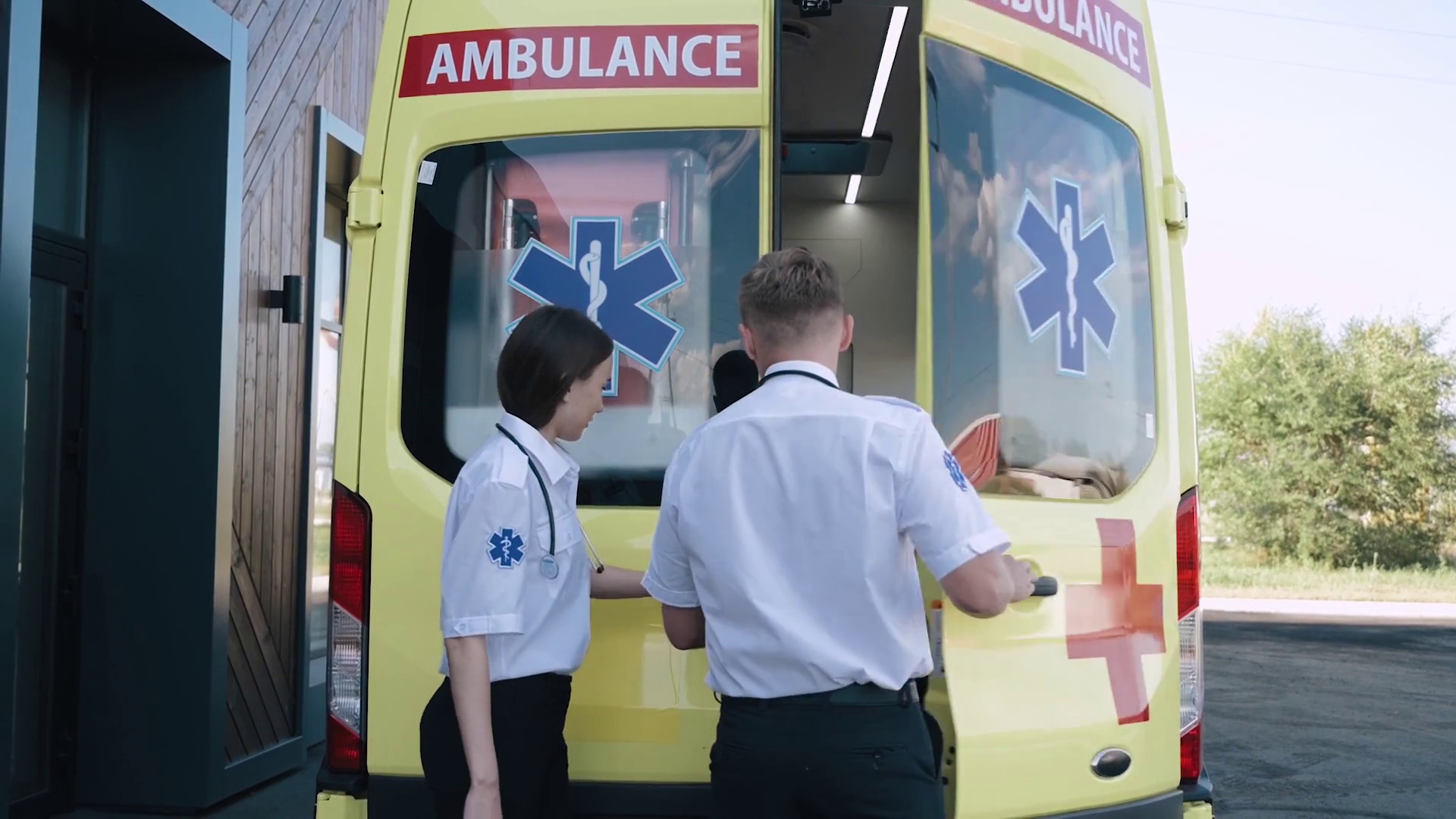} &
        \includegraphics[width=0.165\textwidth, height=0.087\textwidth]{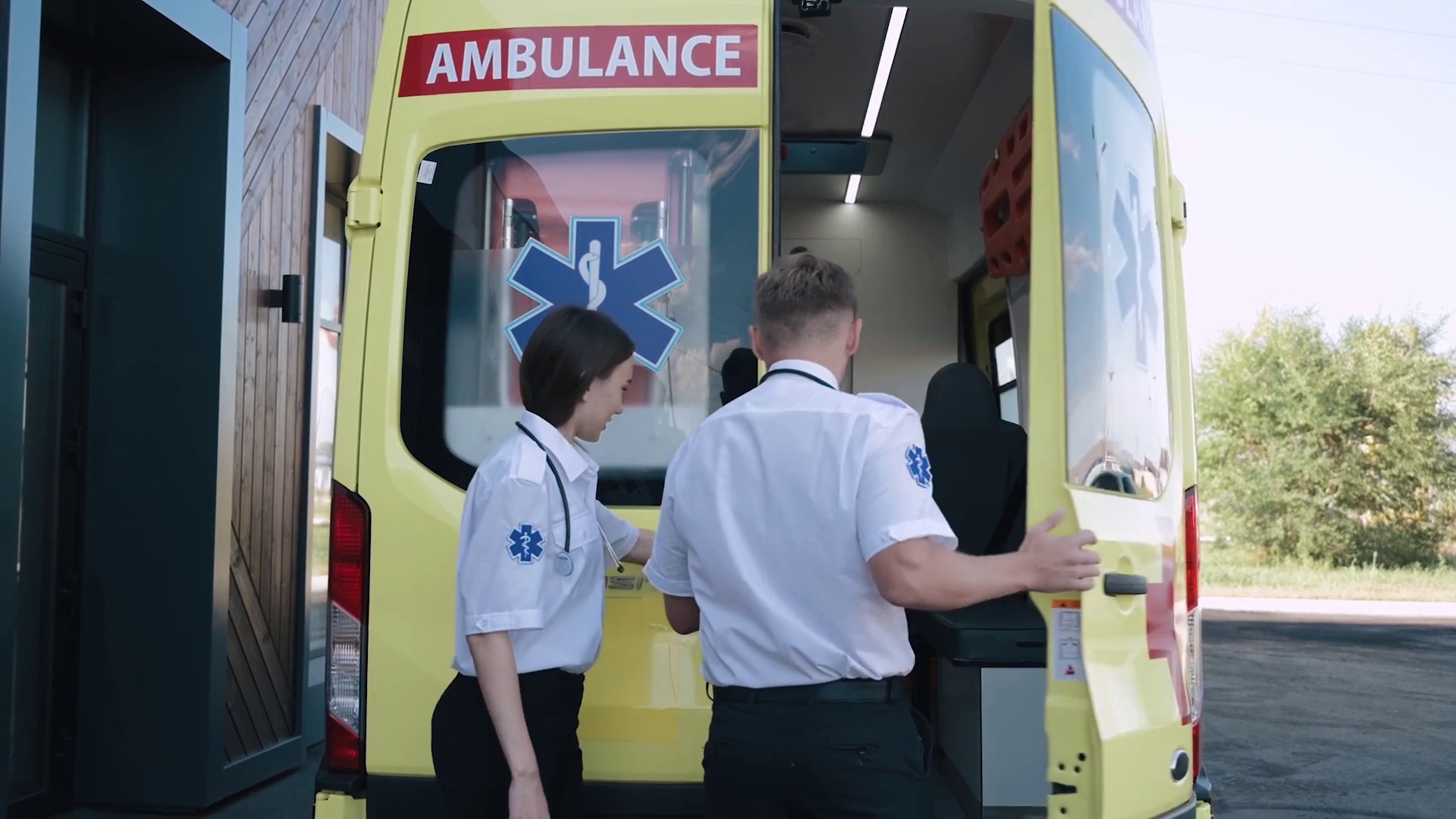} \\

        \multicolumn{3}{c}{Similar motion} && \multicolumn{3}{c}{Similar motion} \\
        \includegraphics[width=0.165\textwidth, height=0.087\textwidth]{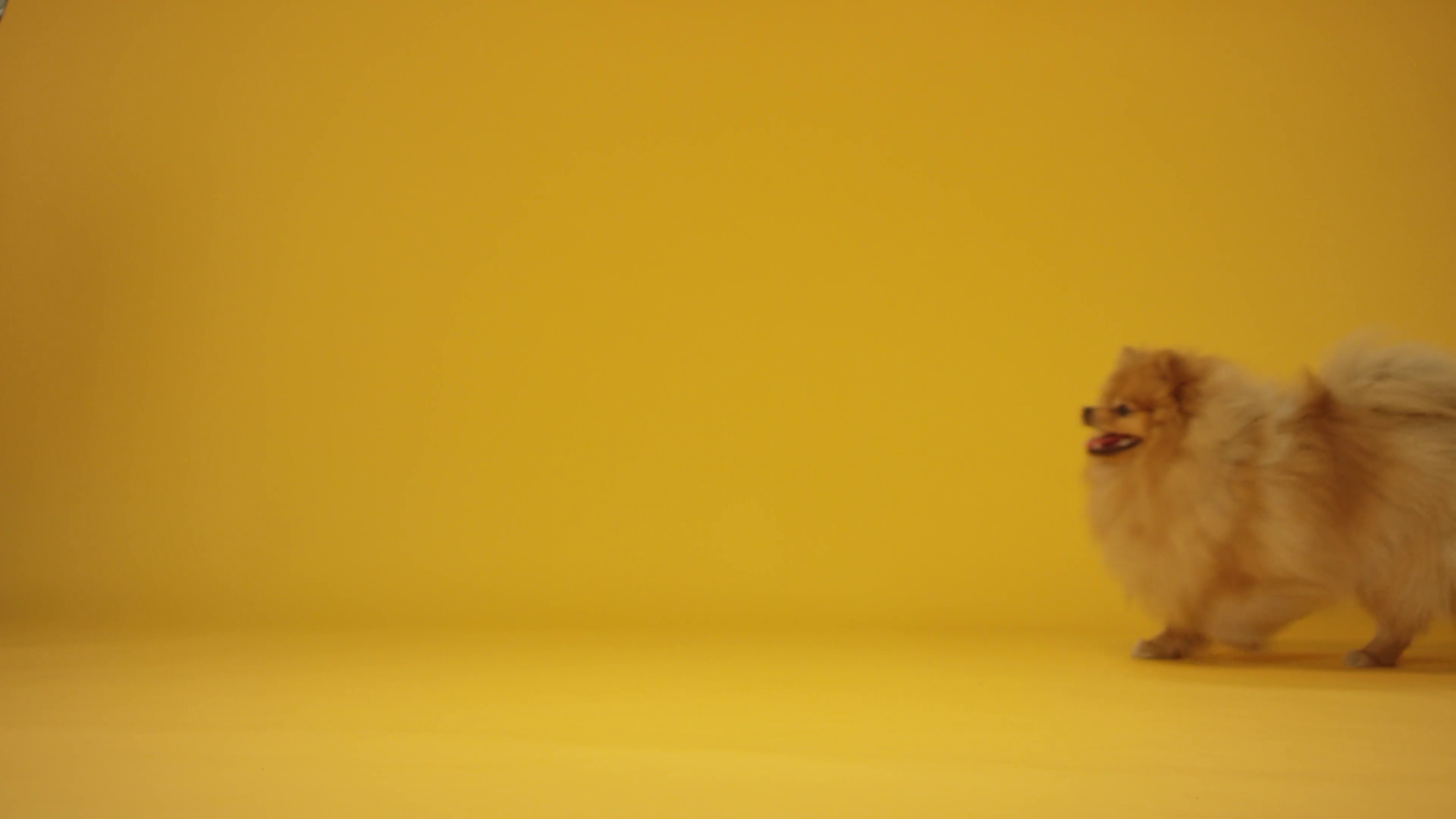} &
        \includegraphics[width=0.165\textwidth, height=0.087\textwidth]{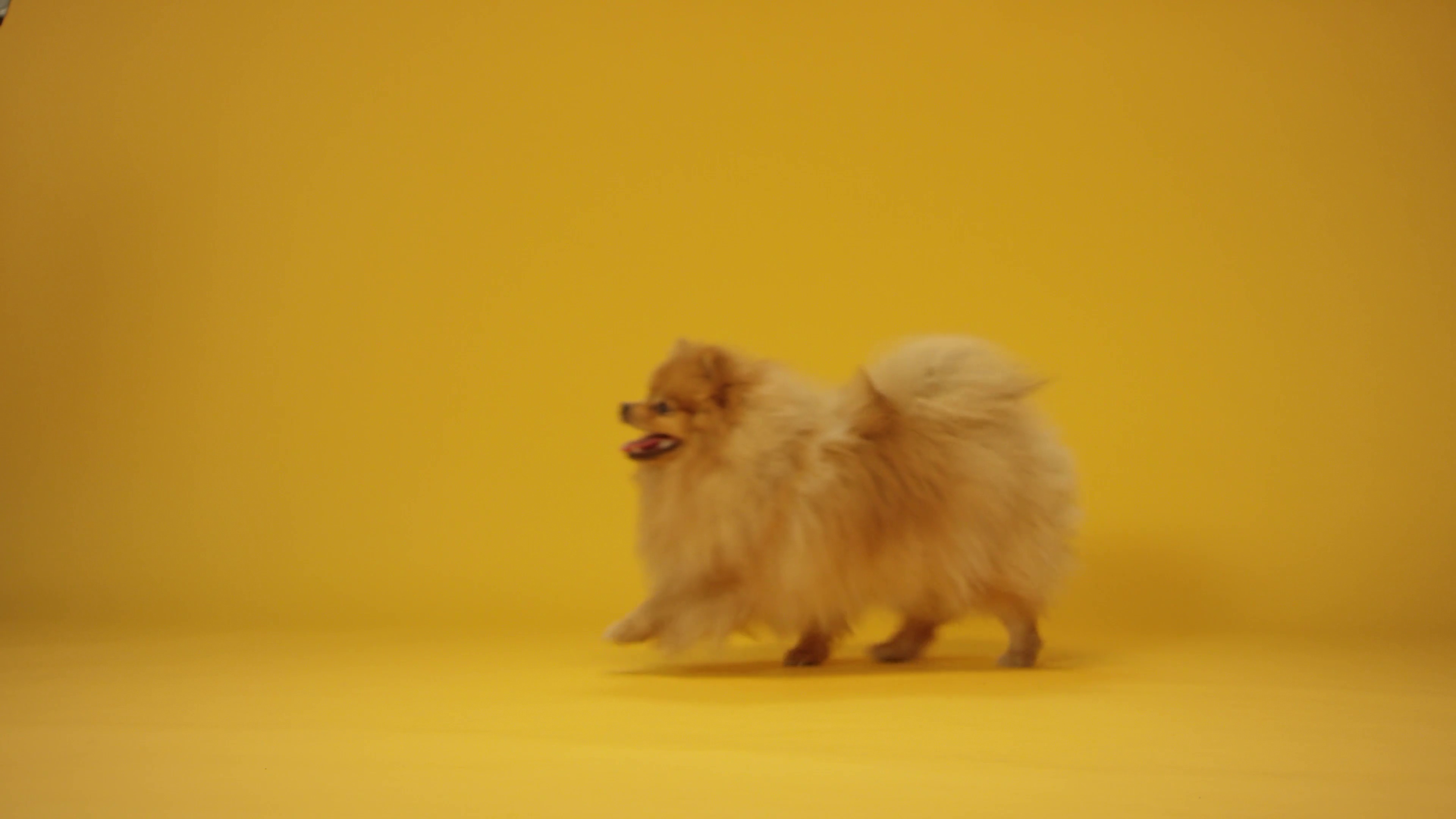} &
        \includegraphics[width=0.165\textwidth, height=0.087\textwidth]{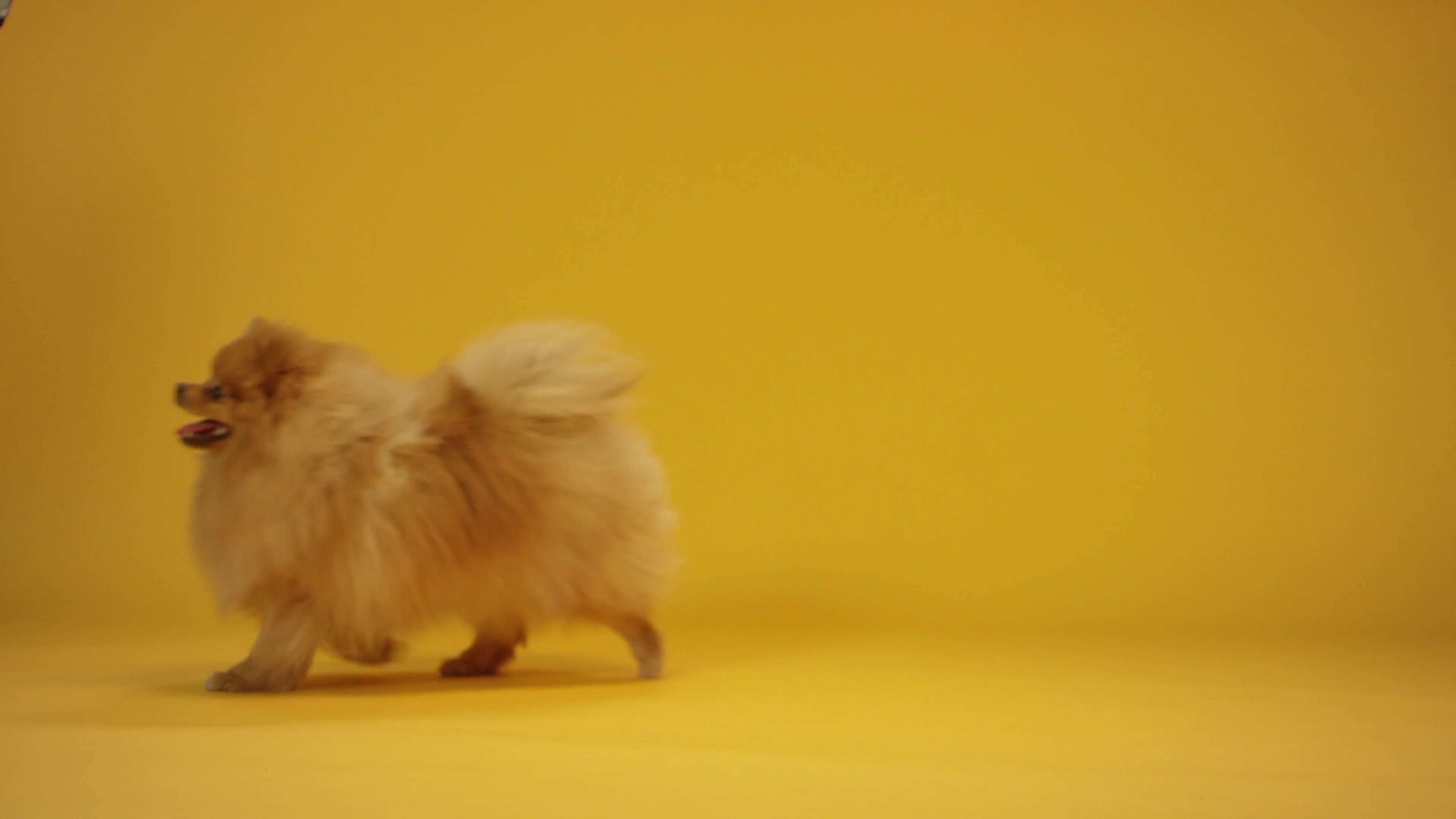} &
        {} &
        \includegraphics[width=0.165\textwidth, height=0.087\textwidth]{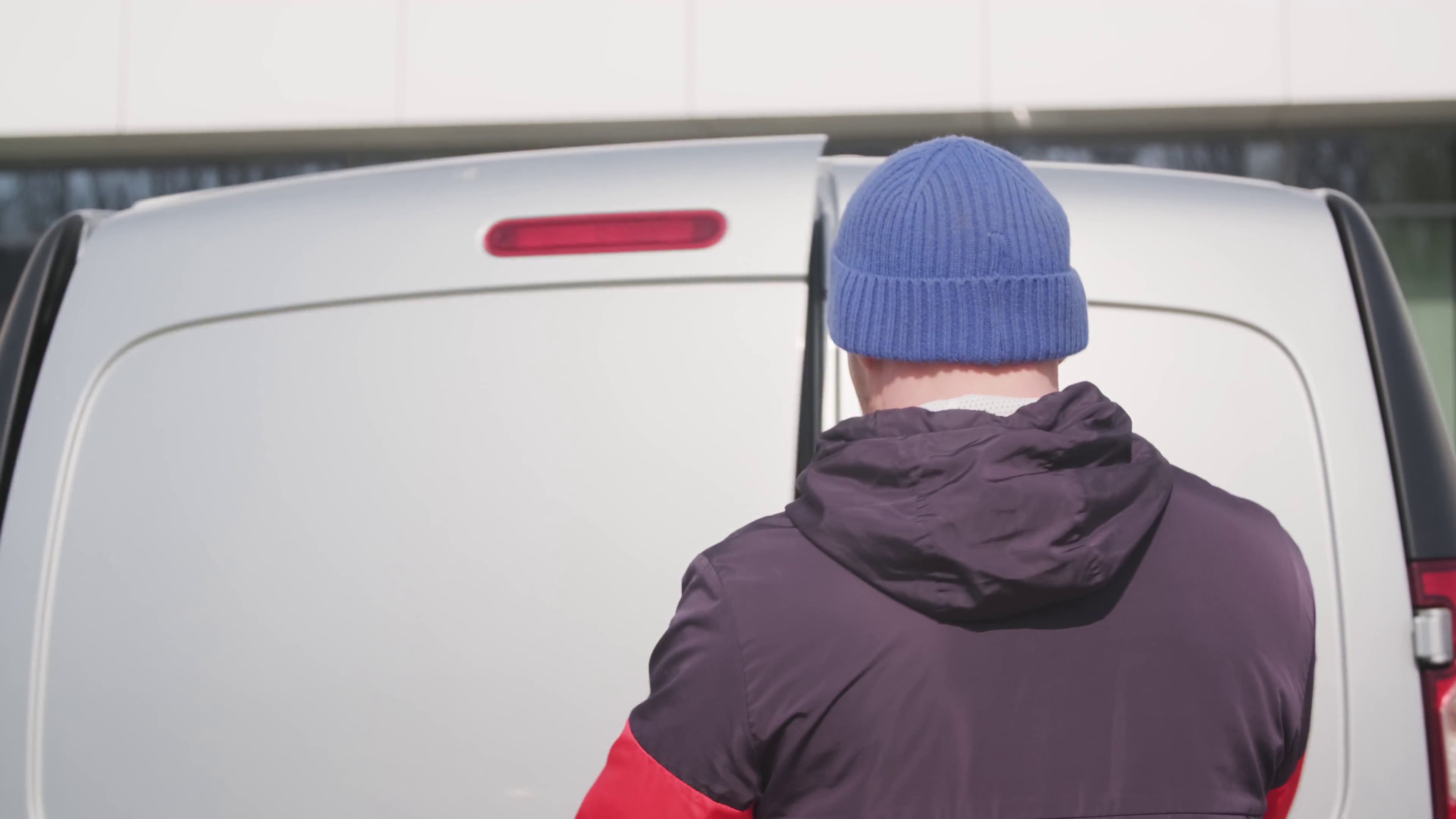} &
        \includegraphics[width=0.165\textwidth, height=0.087\textwidth]{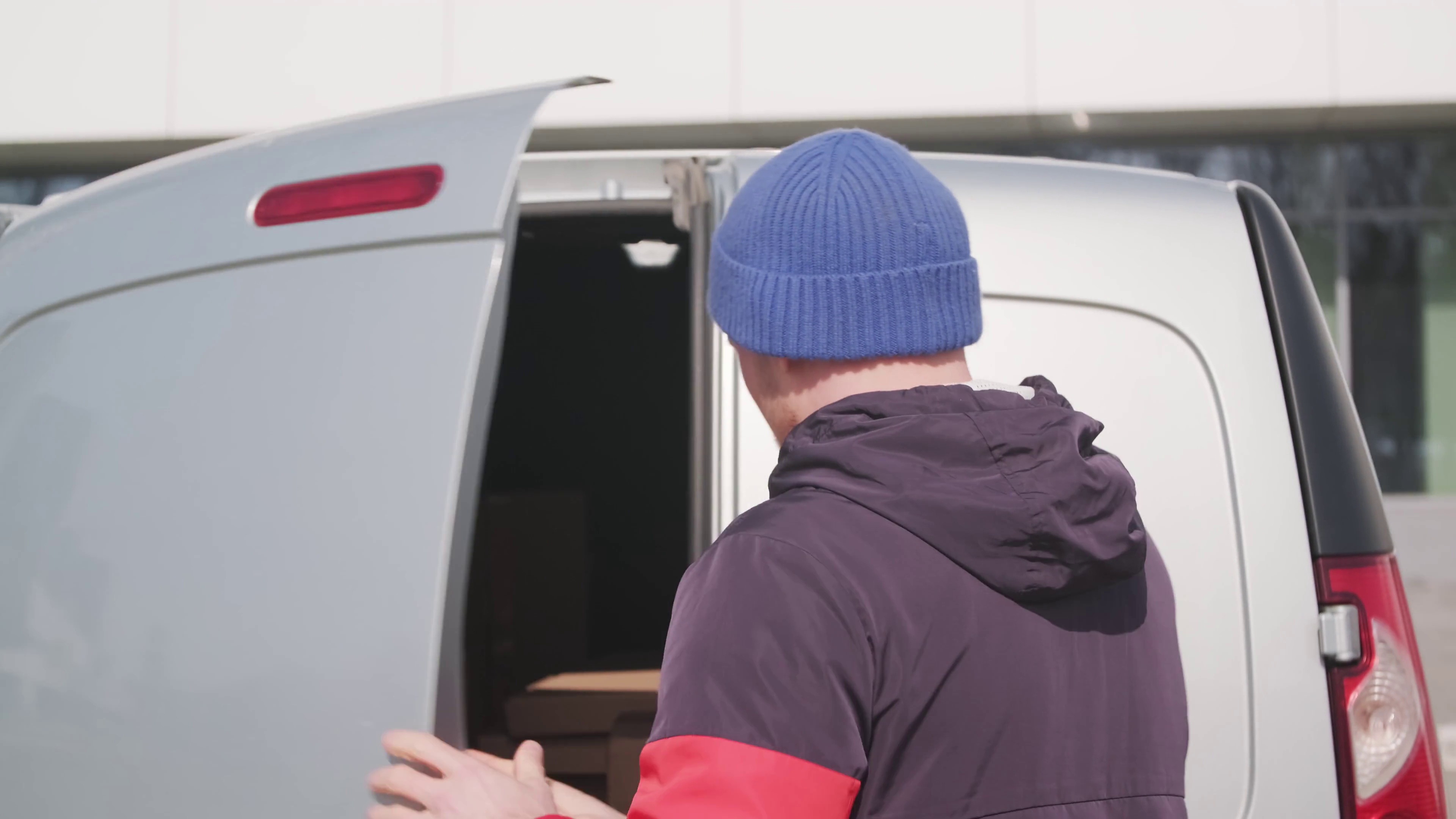} &
        \includegraphics[width=0.165\textwidth, height=0.087\textwidth]{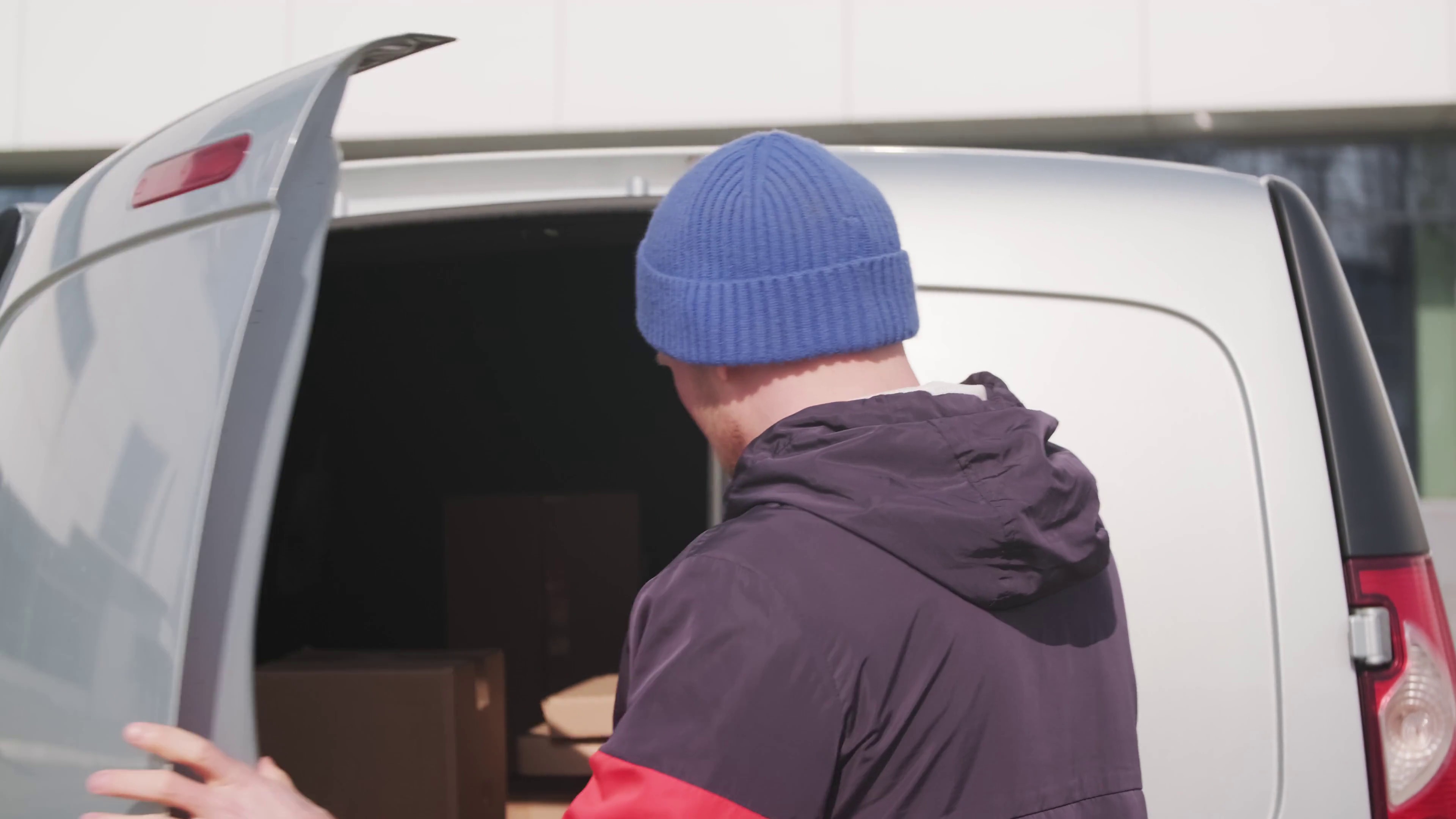} \\

        \multicolumn{3}{c}{Similar appearance} && \multicolumn{3}{c}{Similar appearance} \\
        \includegraphics[width=0.165\textwidth, height=0.087\textwidth]{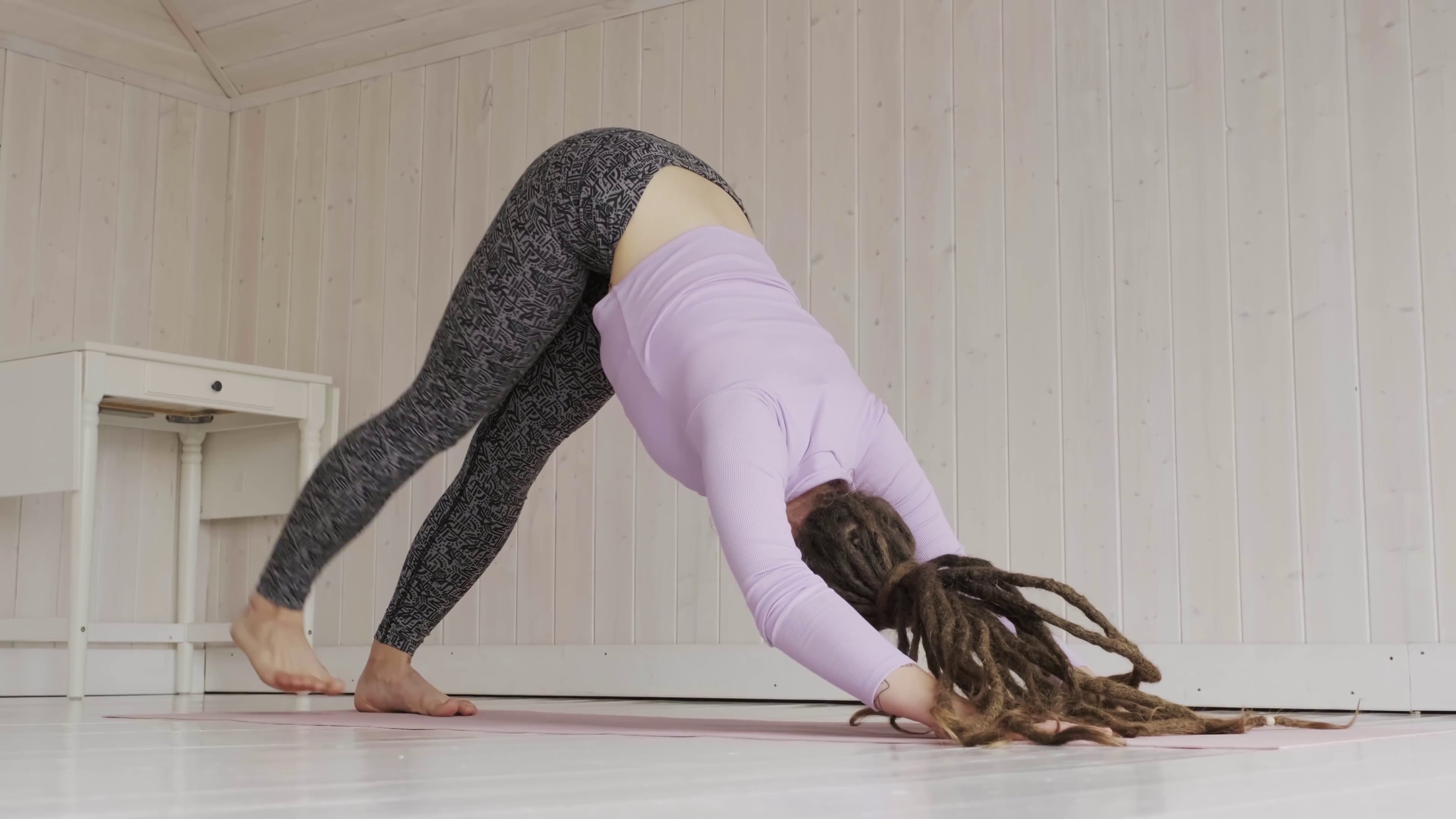} &
        \includegraphics[width=0.165\textwidth, height=0.087\textwidth]{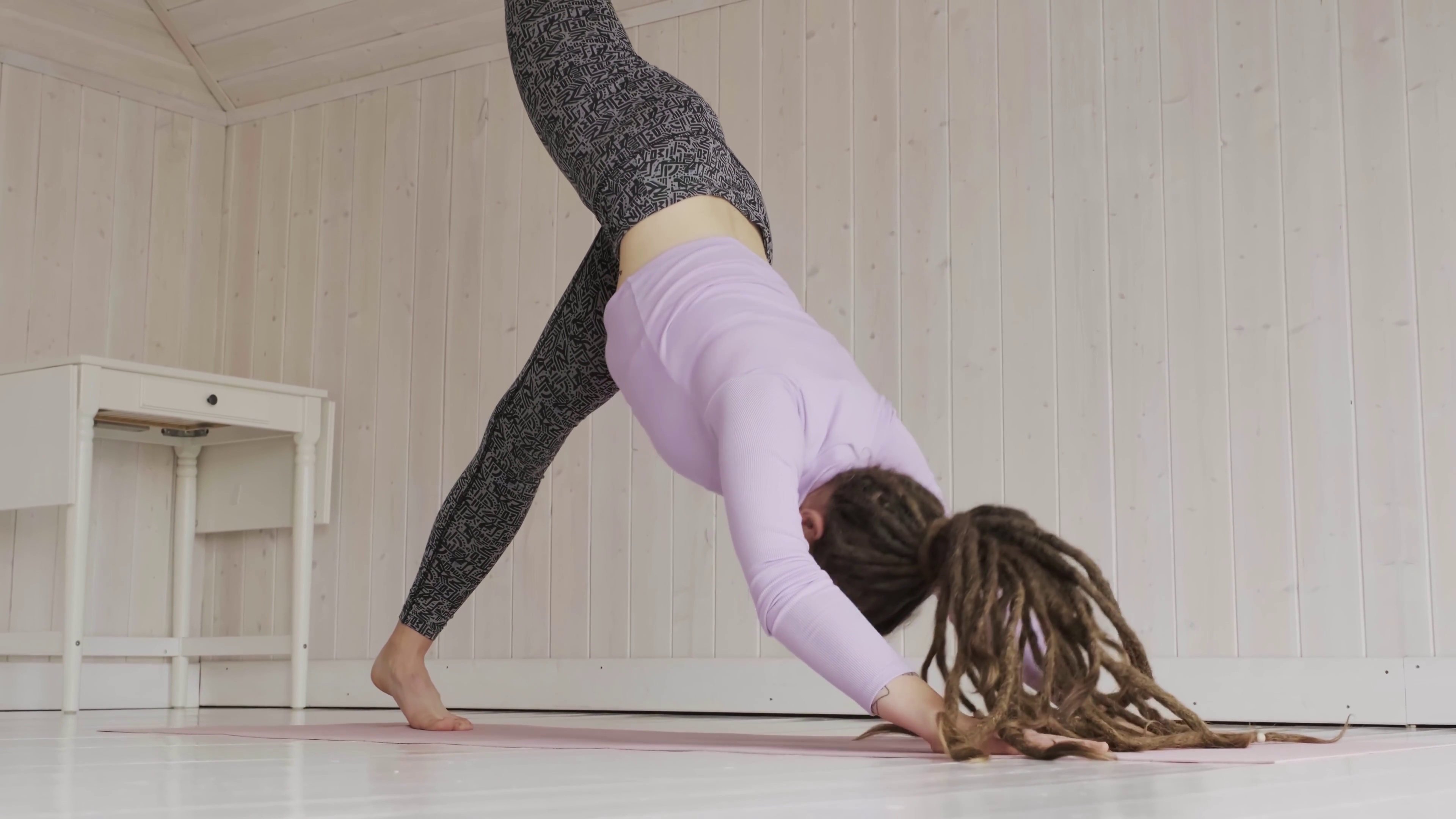} &
        \includegraphics[width=0.165\textwidth, height=0.087\textwidth]{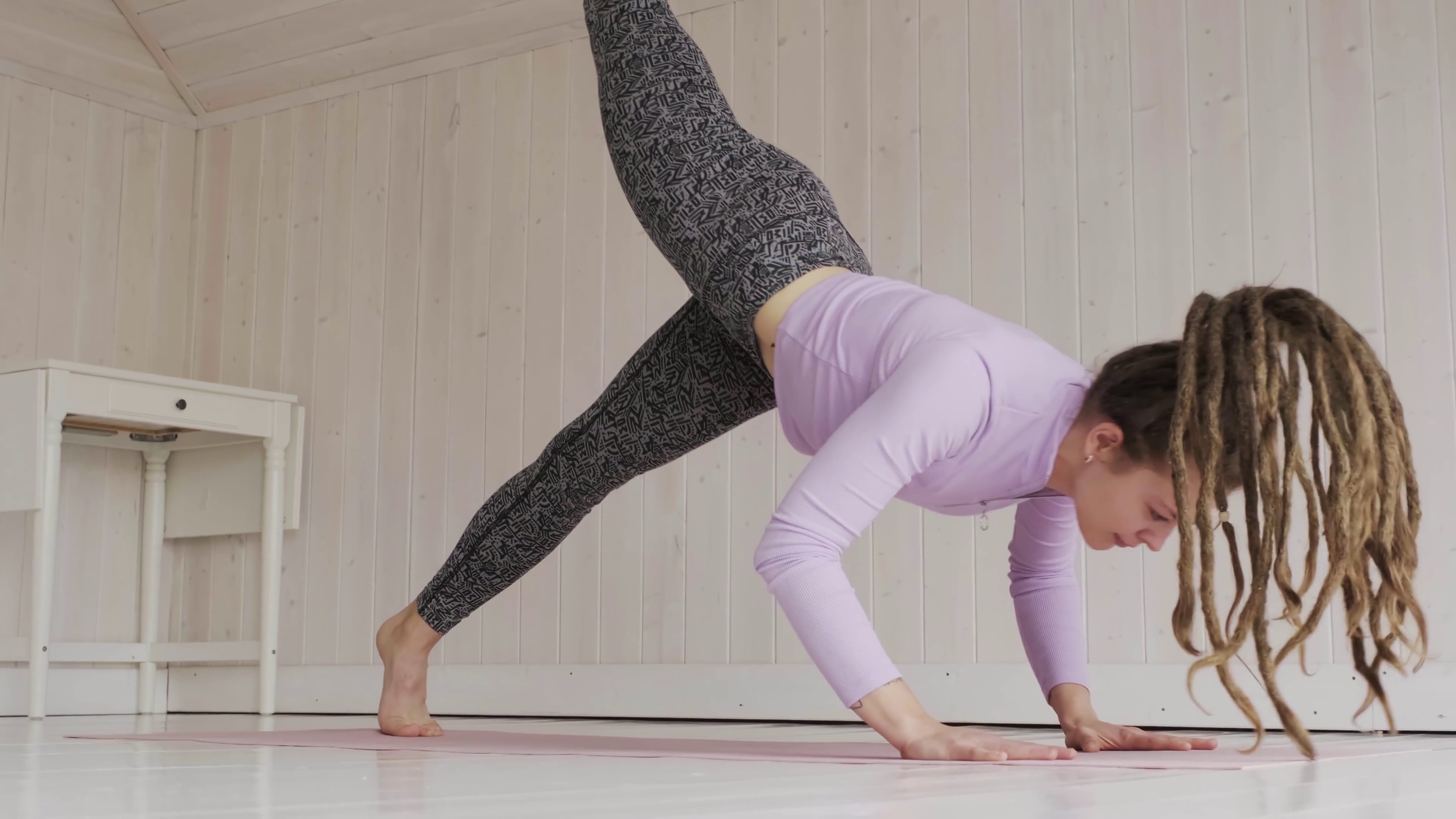} &
        {} &
        \includegraphics[width=0.165\textwidth, height=0.087\textwidth]{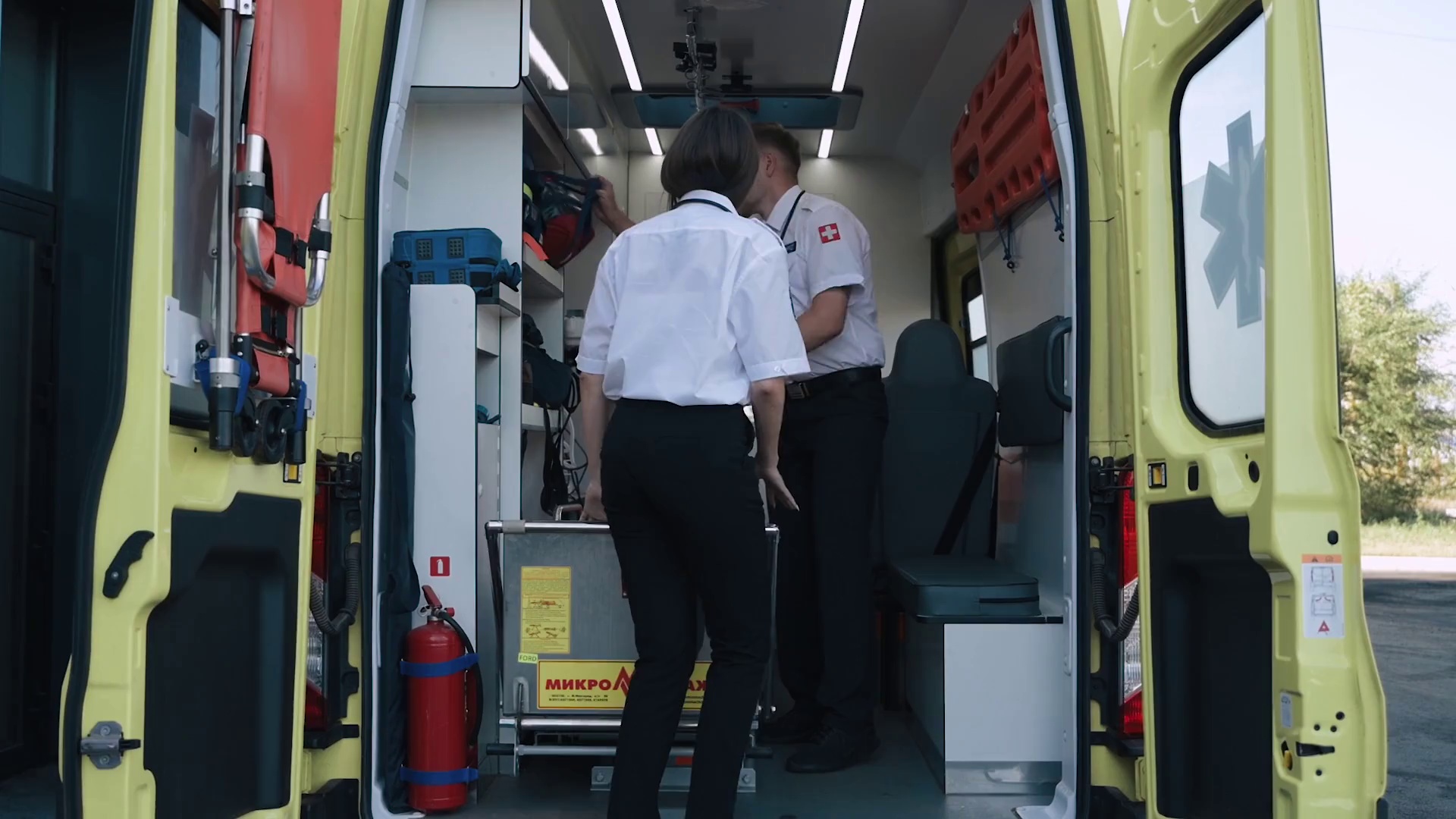} &
        \includegraphics[width=0.165\textwidth, height=0.087\textwidth]{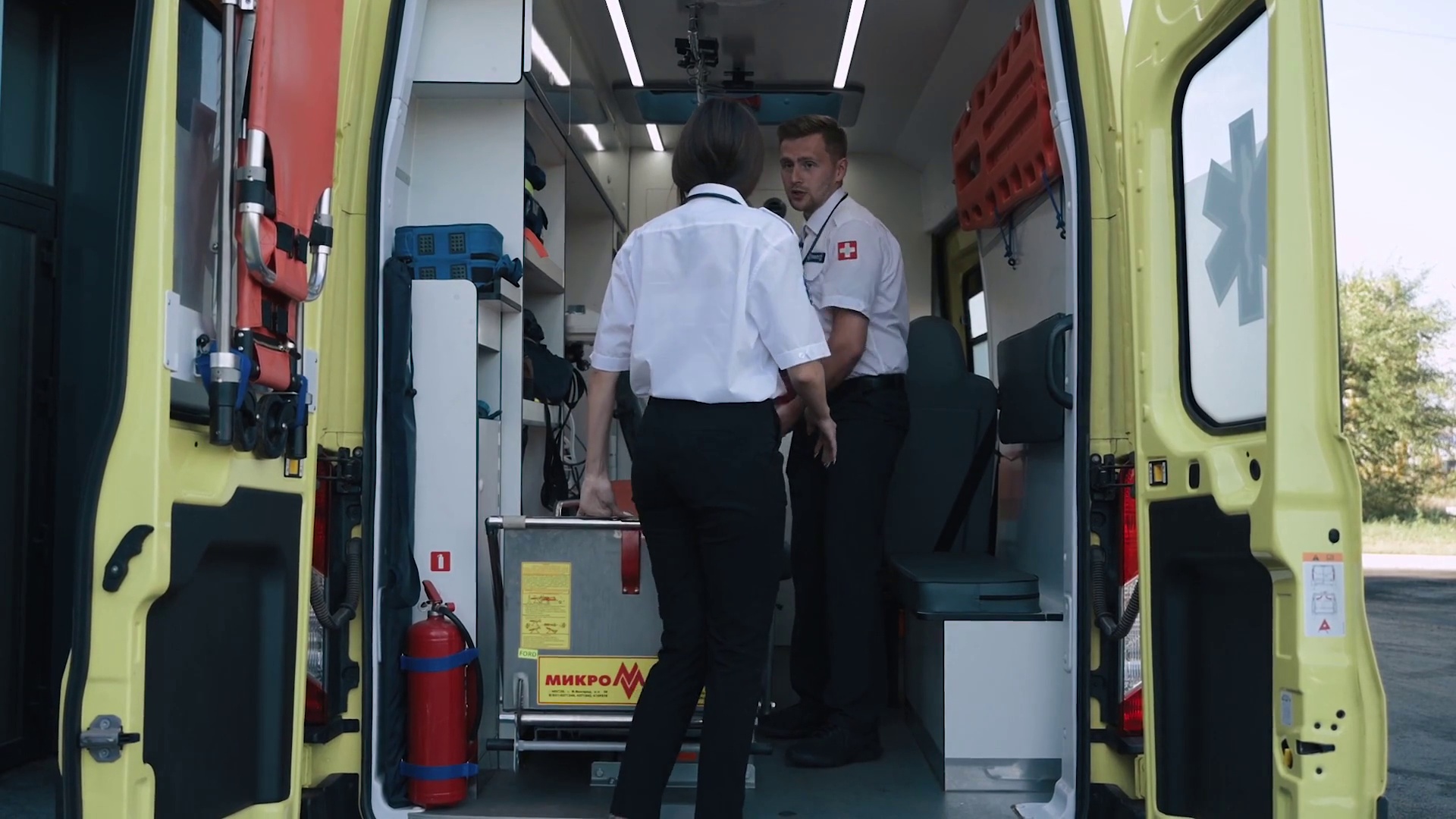} &
        \includegraphics[width=0.165\textwidth, height=0.087\textwidth]{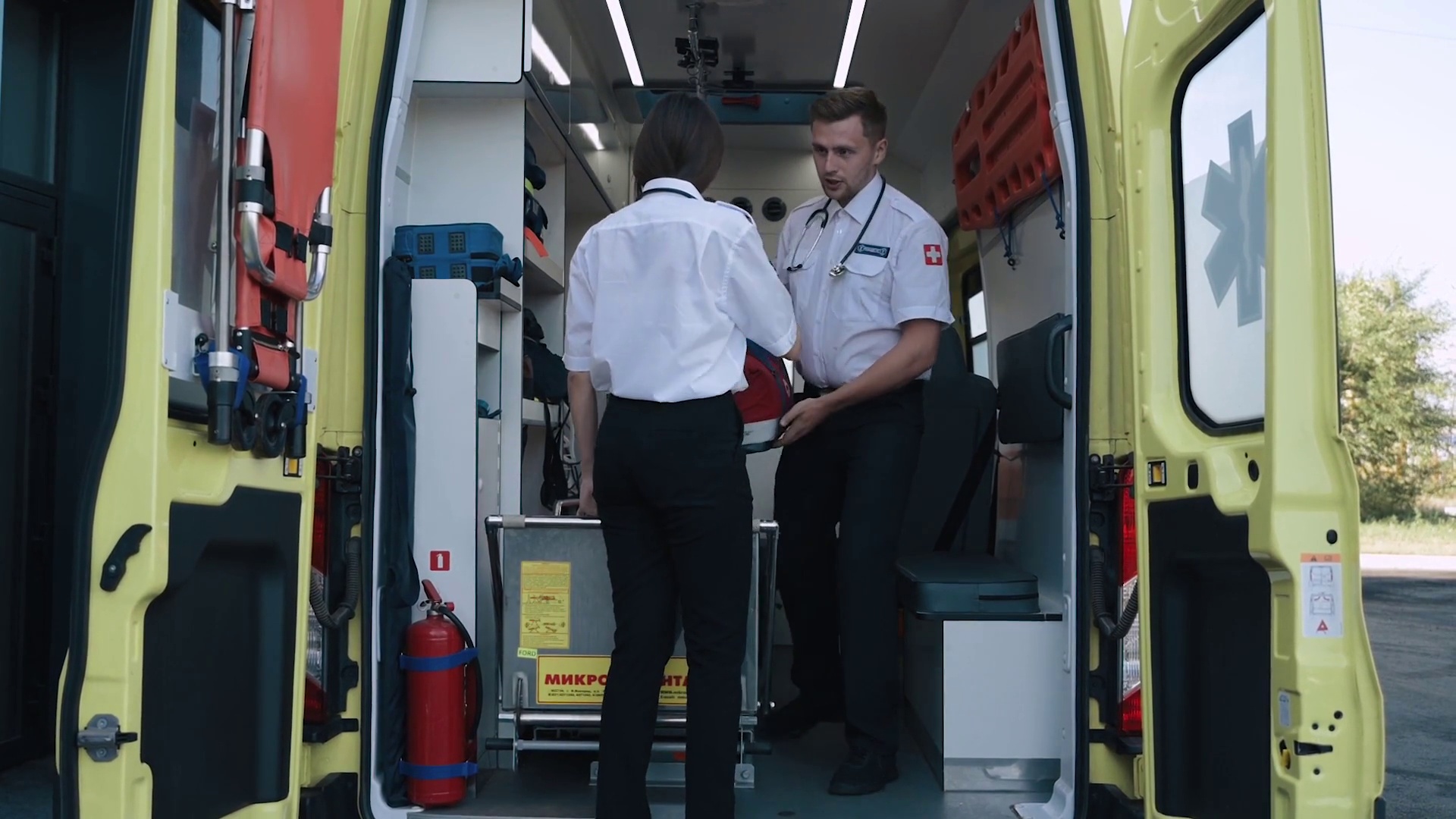} \\

        \multicolumn{3}{c}{(A)} && \multicolumn{3}{c}{(B)} \\
    \end{tabular}
    }
    \caption{
    \textbf{Motion vs. Appearance Bias.} 
    \emph{Left:} the dominant motion is a dog walking; while VideoPrism retrieves a static yoga pose (bottom row) based on the background or inferred label "woman doing yoga", our Semantic Moments (middle row) successfully retrieves a dog walking despite the different background. \emph{Right:} although the motion is opening a door, VideoMAE retrieves a video matching the ``ambulance" context. In contrast, our method aligns with the underlying dynamics, ignoring static appearance or coarse semantics. 
    }
    \vspace{-10pt}
    \label{fig:motionvsapperance}
\end{figure*}

\paragraph{\textbf{Evaluation Protocol}}

Since our primary application is motion-focused \emph{retrieval}, we adopt a retrieval-based evaluation. For each method, we extract video embeddings, $\ell_2$-normalize them, and compute cosine similarity between a query and all candidates. Given a query video, we rank motion-preserving positives and distractors by similarity and report the success rate of closest video retrieval. Because our motion-focused datasets are medium in scale and we emphasize on precision, we highlight this as the most informative metric.

\subsection{Evaluation on \datasetS}
\label{section:quantitative_SimMotion_Synthetic}
We begin with \datasetS, a controlled benchmark that isolates motion similarity under systematic appearance variations. It defines five motion-preserving edit categories (Sec.~\ref{synthetic_dataset}, examples in Fig.~\ref{fig:sync_motions_dataset}) and reports retrieval accuracy. Tab.~\ref{tab:synthetic} summarizes results. 

\begin{table}[t]
\scriptsize
\centering
\caption{\textbf{Synthetic motion-similarity on \datasetS.} Retrieval accuracy (higher is better) across motion-preserving edit categories. The benchmark holds motion fixed while varying appearance factors (object identity/attributes, view, and scene style), exposing where representations over-index on appearance. $\algname$ denotes our moment-based representation instantiated with different frame encoders. Our method achieves the best overall average score.}
\label{tab:synthetic}
\setlength{\tabcolsep}{3pt}
\resizebox{\linewidth}{!}{
\begin{tabular}{lcccccc}
\toprule
& \textbf{Static} & \textbf{Dyn-App} & \textbf{Dyn-Obj} & \textbf{View} & \textbf{Style} & \textbf{Avg} \\
\midrule
CLIP4Clip        & 48.00 & 64.00 & 18.00 & 62.00 & 20.00 & 42.20 \\
X-CLIP           & 68.00 & 80.00 & 12.00 & 64.00 & 32.00 & 51.20 \\
\midrule
SlowFast         & 86.00 & \underline{96.00} & 74.00 & 76.00 & 58.00 & 78.00 \\
TimeSformer      & 76.00 & 86.00 & 36.00 & 62.00 & 32.00 & 58.40 \\
VideoMoCo        & 78.00 & 84.00 & \textbf{82.00} & 46.00 & 32.00 & 64.40 \\
VideoMAE         & 90.00 & \underline{98.00} & 64.00 & \underline{78.00} & 66.00 & 79.20 \\
VideoPrism       & 64.00 & 90.00 & 28.00 & 68.00 & 22.00 & 54.40 \\
V-JEPA2      & 92.00 & 98.00 & 72.00 & 64.00 & 46.00 & 74.40 \\
\midrule
I3D              & \underline{98.00} & \underline{96.00} & 74.00 & 74.00 & \textbf{80.00} & \underline{84.40} \\
CoCLR            & 76.00 & 82.00 & 52.00 & 40.00 & 26.00 & 55.20 \\
MaCLR            & 56.00 & 78.00 & 68.00 & 38.00 & 10.0 & 50.00 \\
\midrule
DINOv2           & 30.00 & 88.00 & 18.00 & 62.00 & 24.00 & 44.40 \\
\midrule
$\text{\algname}_{\text{DINO}}$       & 96.00 & \underline{98.00} & 78.00 & \textbf{82.00} & \underline{78.00} & \textbf{86.40} \\
$\text{\algname}_{\text{VideoPrism}}$ & 94.00 & \textbf{100.00} & 56.00 & 74.00 & 74.00 & 79.60 \\
$\text{\algname}_{\text{VideoMAE}}$   & 92.00 & 94.00 & 76.00 & \textbf{82.00} & 74.00 & 83.60 \\
$\text{\algname}_{\text{V-JEPA2}}$      & \textbf{100.00} & \textbf{100.00} & \underline{80.00} & 70.00 & 72.00 & \underline{84.40} \\
\bottomrule
\end{tabular}
}
\vspace{-0.3cm}
\end{table}

\datasetS ~reveals complementary weaknesses across different baselines. \emph{CLIP-based multimodal} models tend to under-specify motion: when appearance shifts while motion is held fixed, their similarity is unstable. \emph{RGB-supervised} representations are dominated by appearance and are sensitive to style changes. \emph{Optical-flow} models are motion-aware but struggle to generalize across subjects and viewpoints, as flow fields vary with shape and camera geometry even for identical dynamics.

A category-wise analysis clarifies these effects. In \textit{Static Object}, where changes occur only in static regions, CLIP-based and RGB-trained models often collapse when backgrounds differ, whereas flow-based methods work well as static regions contribute little to optical flow. In \textit{Dynamic Appearance}, \textsc{I3D} excels by discounting texture, and \textsc{\algname{}} matches this robustness without explicit flow. In \textit{Dynamic Object} (replacing the moving subject while preserving dynamics), all baselines degrade: flow features drift with subject identity, while \textsc{\algname{}} maintains high similarity by summarizing temporal evolution in a semantics-aware feature space. \textit{View} changes hurt both RGB and flow baselines due to geometric misalignment, whereas \textsc{\algname{}} is comparatively robust by aggregating frame-wise semantics over time rather than relying on correspondence.

Overall, \textsc{\algname{}} achieves the best or competitive performance across categories (Tab.~\ref{tab:synthetic}), with strong gains in \textit{Dynamic Object} and \textit{View}, and near–flow-level robustness in \textit{Static Object} and \textit{Dynamic Appearance}. This indicates that simple moment statistics over semantic features mitigate the primary failure modes exposed by \datasetS{} while remaining training-free and encoder-agnostic.

\subsection{Evaluation on SimMotion-Real}
\label{section:quantitative_SimMotion_Real}
While the synthetic benchmark enables granular, category-level analysis, real-world evaluation is essential: shifts in appearance, camera motion, timing, and scene complexity often break controlled setting assumptions. \textit{SimMotion-Real} comprises unconstrained videos with unsynchronized motion, making it a test of semantic motion similarity rather than geometric correspondence, since actions are related but rarely identical. As shown in Tab.~\ref{tab:real_results}, all methods struggle to get high scores, reflecting real-world variability and noise. Flow-based approaches (e.g., \textsc{I3D}) excel when motion is aligned but lose effectiveness on unsynchronized clips, where flow consistency breaks despite similar semantics; CLIP-based and RGB-only models remain dominated by appearance. In contrast, \textsc{SemanticMoments} maintains strong retrieval accuracy and attains the best overall scores, suggesting that temporal statistics over semantic features are more robust to in-the-wild variability. Despite these gains, the absolute numbers indicate that motion-similarity retrieval in the wild remains a challenging open problem. A visual example is provided in \cref{fig:motionvsapperance}, demonstrating the challenge of separating motion from appearance.

\begin{table}[t]
\centering
\caption{\textbf{Real-world motion retrieval on \datasetR-1K.}
Retrieval accuracy with 1{,}000 candidates per query (one motion-preserving positive). 
$\algname$ denotes our moment-based representation instantiated with different frame encoders. As can be seen, $\algname$ improve the baselines significantly for semantic motion retrieval.}
\label{tab:real_results}
\setlength{\tabcolsep}{6pt}
\begin{tabular}{l c}
\toprule
\textbf{Models} & \textbf{Retrieval Accuracy} \\
\midrule
CLIP4Clip                    & 10.00 \\
X-CLIP                       & 10.00 \\
\midrule
I3D                          & \underline{27.50} \\
CoCLR                        & 5.00  \\
MaCLR                        & 2.50  \\
\midrule
SlowFast                     & 15.00 \\
TimeSformer                  & 12.50 \\
VideoMoCo                    & 12.50 \\
VideoMAE                     & 17.50 \\
VideoPrism                   & 12.50 \\
V-JEPA2                 & \underline{27.50} \\
\midrule
DINOv2                       & 12.50 \\
\midrule
$\text{\algname}_{\text{DINO}}$       & \textbf{42.50} \\
$\text{\algname}_{\text{VideoPrism}}$ & 22.50 \\
$\text{\algname}_{\text{VideoMAE}}$   & 22.50 \\
$\text{\algname}_{\text{V-JEPA2}}$   & \underline{27.50} \\
\bottomrule
\end{tabular}%
\end{table}

\begin{table}[t]
\caption{\textbf{Gesture classification on Jester benchmark.}
Top-1 majority vote and weighted kNN accuracy on the Jester validation set (K=20).
$\algname$ consistently improves performance across different backbones.}
\centering
\footnotesize
\setlength{\tabcolsep}{6pt}
\resizebox{\linewidth}{!}{
\begin{tabular}{lccc}
\toprule
\textbf{Method} &
\textbf{Acc@1}$_{\text{maj}}$ &
\textbf{Acc@1}$_{\text{w-kNN}}$ &
\textbf{Acc@5}$_{\text{w-kNN}}$ \\
\midrule

X-CLIP & 26.0 & 20.2 & 42.2 \\
Clip4CLIP & 8.5 & 8.3 & 11.3 \\
TimeSFormer & 16.7 & 10.3 & 20.1 \\
SlowFast & 19.7 & 17.2 & 34.8 \\
VideoMoCo & 12.1 & 12.0 & 24.4 \\
I3D & 26.8 & 25.8 & \underline{53.2} \\
VideoMAE & 23.8 & 22.7 & 43.4 \\
V-JEPA2 & 12.5 & 12.3 & 20.7 \\
DINOv2 & 8.5 & 7.8 & 9.7 \\
\midrule
$\text{\algname}_{\text{DINO}}$ & 25.7 & 25.0 & 50.8 \\
$\text{\algname}_{\text{VideoMAE}}$  & \underline{26.9} & \underline{26.5} & \textbf{55.0} \\
$\text{\algname}_{\text{V-JEPA2}}$ & \textbf{28.6} & \textbf{28.3} & 47.2 \\

\bottomrule

\end{tabular}
}
\label{tab:jester_knn}
\end{table}

\subsection{Gesture-Level Evaluation on Jester}

We further extend our evaluation to the publicly available Jester gesture
benchmark~\cite{materzynska2019jester}, which contains videos annotated with
distinct gesture motion categories.

We evaluate whether SemanticMoments improves gesture-level separability in
the embedding space under different video representations.
To quantify this effect without training an additional classifier, we adopt
a standard kNN evaluation protocol (K=20) on the validation set.
For each query video, we retrieve its $K$ nearest neighbors and predict the
gesture label using their annotations.
Majority-vote accuracy assigns the most frequent neighbor label, while
weighted kNN additionally weights each neighbor contribution by its
similarity to the query.

As shown in Table~\ref{tab:jester_knn}, applying \algname\ consistently improves the metrics across all backbones, indicating stronger motion
representations.

\begin{table}[t]
\centering
\caption{\textbf{Ablation on SimMotion-Real.}
We systematically analyze our DINO-based moment representation across three complementary axes:
(1) \emph{Moment configuration} — comparing single-moment versus multi-moment setups to assess how temporal coverage influences motion alignment.
(2) \emph{Representation level} — applying moments at different abstraction levels, including frame-level (global embeddings), patch-level (spatially localized features), and patch-difference (temporal gradients between patches), to isolate the contribution of spatial granularity and motion sensitivity.
(3) \emph{Embedding combination} — evaluating how final representations are merged, either by simple \emph{summation} or \emph{concatenation}, to study the effect of interaction strength between moment features.
Together, these experiments disentangle how moment granularity, representation hierarchy, and feature fusion each contribute to accurate motion-centric retrieval.}
\label{tab:ablations}

\setlength{\tabcolsep}{6pt}
\begin{tabular}{l c}
\toprule
\textbf{Configuration} & \textbf{Retrieval Accuracy}\\
\midrule
$\text{DINO}_{(1, 0, 0)\text{-}patch\text{-}concat}$      & 12.50 \\
$\text{DINO}_{(0, 1, 0)\text{-}patch\text{-}concat}$      & 10.00 \\
$\text{DINO}_{(1, 1, 0)\text{-}patch\text{-}concat}$      & 12.50 \\
$\text{DINO}_{(1, 8, 0)\text{-}patch\text{-}concat}$      & 15.00 \\
$\text{DINO}_{(1, 1, 1)\text{-}patch\text{-}concat}$      & \underline{30.00}\\
\midrule
$\text{DINO}_{(1, 8, 4)\text{-}patch\text{-}sum}$         & \underline{30.00} \\
$\text{DINO}_{(1, 8, 4)\text{-}frame\text{-}concat}$      & 12.50 \\
$\text{DINO}_{(1, 8, 4)\text{-}diff\text{-}patch\text{-}concat}$ & 22.50 \\
\midrule
$\text{DINO}_{(1, 8, 4)\text{-}patch\text{-}concat}$      & \textbf{42.50} \\
\bottomrule
\end{tabular}%
\vspace{-0.3cm}
\end{table}
\begin{table}[t]
  \centering
  \caption{\textbf{Effect of temporal sampling on SimMotion-Real retrieval.}
Retrieval accuracy with different numbers of uniformly sampled frames per video. Best results are achieved at 32 frames.
  }
  \label{tab:frames}
  \setlength{\tabcolsep}{3pt}
  \begin{tabular}{lccccc}
    \toprule
    \textbf{number of frames} & \textbf{4} & \textbf{8} & \textbf{16} & \textbf{32} & \textbf{64}\\
    \midrule
    \textbf{Retrieval Accuracy} & 22.5 & 35 & 40 & 42.5 & 35 \\
    \bottomrule
  \end{tabular}
\end{table}

\subsection{Ablation Studies}
\label{section:ablation_studies}
We evaluate \algname\ on \textit{SimMotion-Real} by varying three design axes: the order and weighting of temporal moments, the representation level at which moments operate (frame, patch, or patch-difference), and the fusion strategy used to combine moment embeddings. As summarized in Tab.~\ref{tab:ablations}, incorporating higher-order moments consistently improves motion alignment over single-moment baselines, indicating that richer temporal statistics capture complementary dynamics beyond average trends.

Tab.~\ref{tab:ablations} further shows that operating at localized, patch-level granularity better preserves fine motion structure than global frame-level representations, and that applying moments directly on raw patch embeddings outperforms applying them to patch-difference representations. For fusion, additive integration provides strong precision with a compact representation, while concatenation can favor broader recall at the cost of higher dimensionality. Overall, the ablations support our design choice of multi-order localized moment modeling.
Finally, Tab.~\ref{tab:frames} analyzes the effect of temporal
sampling density, showing that retrieval performance improves up to 32 uniformly
sampled frames, after which gains saturate.

\subsection{Limitations}
\label{section:limitations}

While our method performs well in most cases, motion-similarity retrieval in the wild remains challenging. First, some motions are inherently difficult (e.g., fine hand gestures, long-horizon actions, multi-agent interactions). Being training-free, our approach cannot be tuned to these corner cases as effectively as methods finetuned on designated datasets --- an avenue for future work. Second, the field lacks a strong, universal video representation comparable to CLIP/DINO for images. Since our method depends on such backbones, its ceiling is bounded by their quality, and we expect video-native backbones to yield significant gains. Finally, failures persist for extremely subtle dynamics (e.g., breathing) and for motions defined by the absence of motion (e.g., waiting). Overall, we view this work as a step toward motion-centric video understanding, and anticipate that improved video backbones and targeted training will help close these gaps.

\vspace{-3pt}
\section{Conclusion}
We introduce the task of \emph{motion-centric video similarity}, targeting how well representations capture and compare motion. Existing retrieval benchmarks are suboptimal for this goal, as labels are often recoverable from static appearance or scene context rather than dynamics. To address this, we propose two dedicated evaluations: \emph{SimMotion-Synthetic}, a controlled, diagnostic benchmark, and \emph{SimMotion-Real}, an unconstrained, real-world benchmark. Together, they form a focused testbed for analyzing motion perception in video representations and reveal systematic limitations of current models. Building on these insights, we present \emph{\algname{}}, a training-free representation that encodes motion via temporal statistics of pretrained semantic features. Despite its simplicity, \algname{} achieves strong motion alignment across multiple backbones and consistently outperforms prior approaches. Still, results on real-world data indicate that models remain far from human-level motion perception, underscoring both the challenge and opportunity of this task. We position SimMotion and \algname as an initial but foundational step toward robust, motion-aware, and perceptually aligned video representations.

{
    \small
    \bibliographystyle{ieeenat_fullname}
    \bibliography{main}
}


\end{document}